%% file: main.tex
\documentclass[10pt,twocolumn,letterpaper]{article}

\usepackage[pagenumbers]{iccv} 


\usepackage[pagebackref,breaklinks,colorlinks,allcolors=iccvblue]{hyperref}
\usepackage{booktabs}
\usepackage{colortbl}
\usepackage{makecell}

\usepackage[table]{xcolor} 
\definecolor{iccvblue}{rgb}{0.21,0.49,0.74}

\usepackage{multirow}
\usepackage{makecell}
\newcommand{\partitle}[1]{\smallskip \noindent \textbf{#1}.}

\def\paperID{2917} 
\def\confName{ICCV}
\def\confYear{2025}

\title{MaskHand: Generative Masked Modeling for Robust Hand Mesh Reconstruction in the Wild}

\author{
Muhammad Usama Saleem, 
Ekkasit Pinyoanuntapong, 
Mayur Jagdishbhai Patel, 
Hongfei Xue, \\ 
Ahmed Helmy, 
Srijan Das, 
Pu Wang\\
University of North Carolina at Charlotte, Charlotte, NC, USA\\
{\tt\small \{msaleem2, epinyoan, mpate169, hxue2, ahelmy1, sdas24, pwang13\}@charlotte.edu}
}

\begin{document}

\makeatletter
\let\@oldmaketitle\@maketitle 
\renewcommand{\@maketitle}{
  \@oldmaketitle 
  \vspace{1em} 
  \centering
  \includegraphics[width=0.9\linewidth]{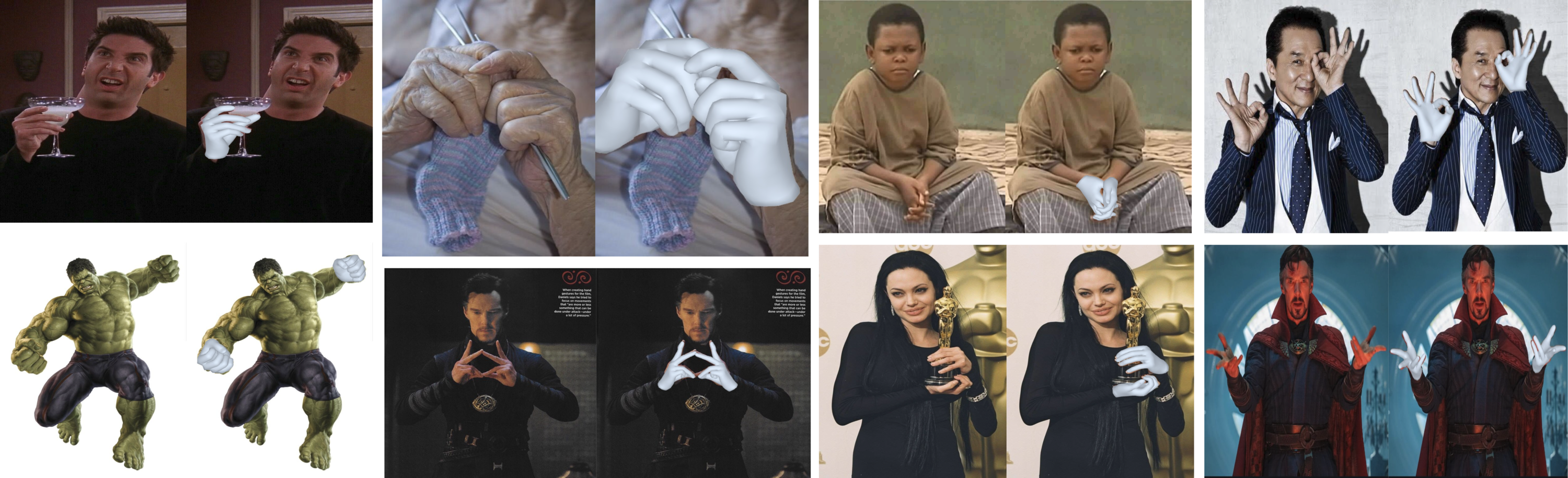}
  \captionof{figure}{MaskHand: a novel generative masked model for accurate and robust 3D hand mesh recovery from single RGB images, excelling in diverse scenarios like occlusions, hand-object interactions, and varied appearances. Watch our Supplemental Video to see it in action!}
  \label{fig:MMHMR_landing}

  \bigskip
}
\makeatother

\makeatother

\maketitle

\input{sec/0_abstract}    
\input{sec/1_intro}
\input{sec/2_related}
\input{sec/3_method}

\input{sec/4_experiment}
\input{sec/5_ablation_study}

\input{sec/6_conclusion}
{
    \small
    \bibliographystyle{ieeenat_fullname}
    \bibliography{main}
}
\input{sec/X_suppl}

\end{document}

%% file: sec/0_abstract.tex
\begin{abstract}

Reconstructing a 3D hand mesh from a single RGB image is challenging due to complex articulations, self-occlusions, and depth ambiguities. Traditional discriminative methods, which learn a deterministic mapping from a 2D image to a single 3D mesh, often struggle with the inherent ambiguities in 2D-to-3D mapping. To address this challenge, we propose \textbf{\textit{MaskHand}}, a novel generative masked model for hand mesh recovery that synthesizes plausible 3D hand meshes by learning and sampling from the probabilistic distribution of the ambiguous 2D-to-3D mapping process. MaskHand consists of two key components: (1) a \textbf{\textit{VQ-MANO}}, which encodes 3D hand articulations as discrete pose tokens in a latent space, and (2) a \textbf{\textit{Context-Guided Masked Transformer}} that randomly masks out pose tokens and learns their joint distribution, conditioned on corrupted token sequence, image context, and 2D pose cues. This learned distribution facilitates confidence-guided sampling during inference, producing mesh reconstructions with low uncertainty and high precision. Extensive evaluations on benchmark and real-world datasets demonstrate that MaskHand achieves state-of-the-art accuracy, robustness, and realism in 3D hand mesh reconstruction. Project website: \url{https://m-usamasaleem.github.io/publication/MaskHand/MaskHand.html}.

\end{abstract}


%% file: sec/1_intro.tex
\vspace{-15pt}
\section{Introduction}
\label{sec:intro}
Hand mesh recovery has gained significant interest in computer vision due to its broad applications in fields such as robotics, human-computer interaction \cite{gesture_keyboard, mid_air}, animation, and AR/VR \cite{handavatar, 3DHumanPoseEstimationVideo}. While previous methods have explored markerless, image-based hand understanding, most depend on depth cameras \cite{baek2018augmented, garcia2018first, moon2018v2v, qian2014realtime, sridhar2015fast} or multi-view images \cite{ballan2012motion, gomez2017large, simon2017hand, sridhar2013interactive}. Consequently, most of these methods are not feasible for real-world applications where only monocular RGB images are accessible. On the other hand, monocular hand mesh recovery from a single RGB image, especially without body context or explicit camera parameters, is highly challenging due to substantial variations in hand appearance in 3D space, frequent self-occlusions, and complex articulations.

Recent advances, especially in transformer-based methods, have shown significant promise in monocular hand mesh recovery (HMR) by capturing intricate hand structures and spatial relationships. For instance, METRO \cite{cho2022FastMETRO} and MeshGraphormer \cite{lin2021mesh} utilize multi-layer attention mechanisms to model both vertex-vertex and vertex-joint interactions, thereby enhancing mesh fidelity. Later, HaMeR \cite{pavlakos2024reconstructing} illustrated the scaling benefits of large vision transformers and extensive datasets for HMR, achieving improved reconstruction accuracy. However, these methods are inherently discriminative, producing deterministic outputs for each image. Consequently, they face limitations in complex, real-world scenes where ambiguities arise due to occlusions, hand-object interactions, and challenging viewpoints are prevalent.
\begin{figure}[ht] 
    \centering
    \includegraphics[width=1\linewidth]{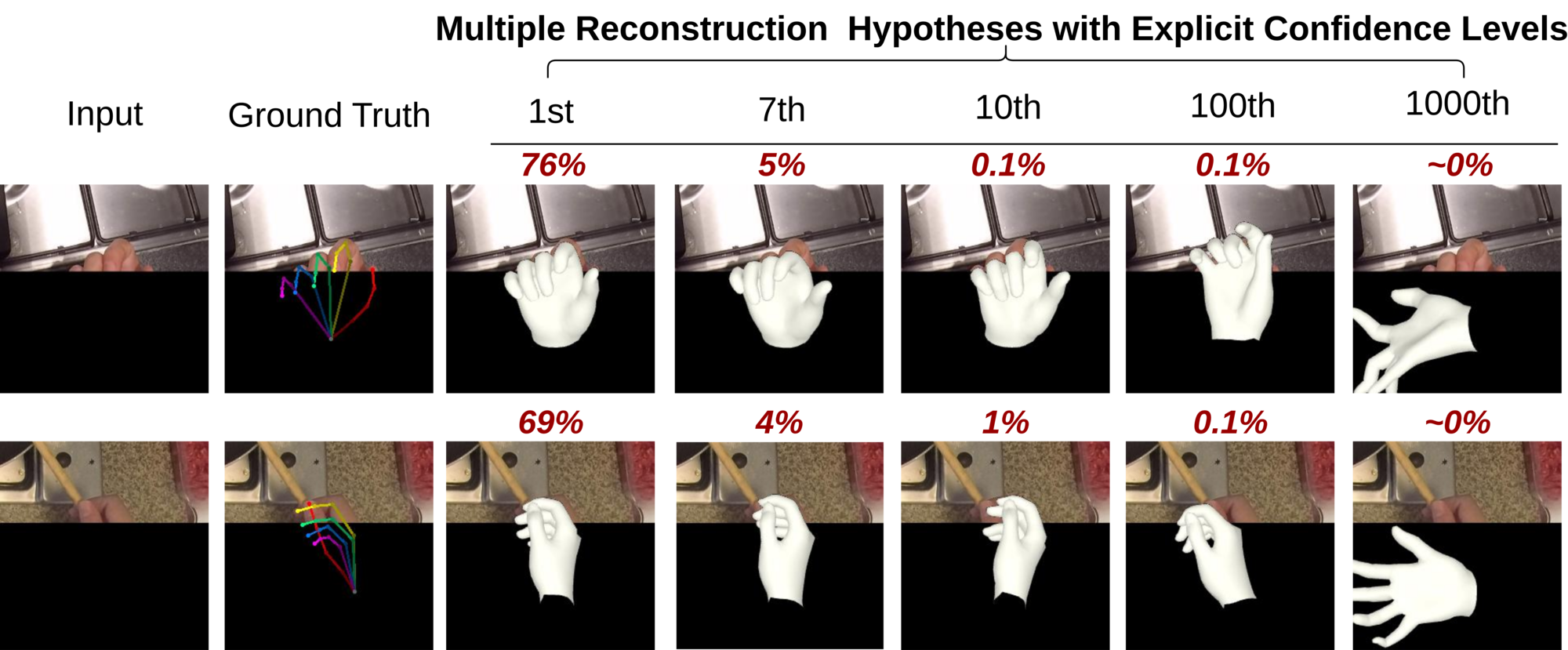}
    \caption{MaskHand is a quantifiable probabilistic HMR method that can learn 2D-to-3D mapping distributions and explicitly estimate confidence levels or prediction probabilities for all mesh reconstruction hypotheses  }
    \label{fig:MaskHand_conf_main} 
\end{figure}

To overcome these limitations, we introduce MaskHand, a novel generative masked model designed for accurate 3D hand mesh recovery. By learning and sampling from the underlying joint distribution of hand articulations and image features, our model synthesizes most probable 3D hand meshes, mitigating ambiguities inherent in single-view reconstruction as shown in Figure \ref{fig:MaskHand_conf_main}. MaskHand consists of two main components: a VQ-MANO and a context-guided masked transformer, which are trained in two consecutive stages. In the first stage, VQ-MANO is trained with Vector Quantized Variational Autoencoders (VQ-VAE) \cite{van2017neural} to encode continuous hand poses (e.g., joint rotations) of the MONO parametric hand model into  a sequence of discrete pose tokens. The discretization in VQ-MANO is crucial for generative masked modeling, allowing the model to learn per-token probabilities conditioned on the input image. In the second stage, the token sequence is partially masked and the context-guided masked transformer is trained to reconstruct the masked tokens by learning token conditional distribution, based on multiple contextual clues, including corrupted token sequence, image features, and 2D pose structures.  The learned distribution enables uncertainty quantification at each prediction step, facilitating confidence-guided sampling to iteratively refine mesh reconstructions. By selectively retaining high-confidence tokens during inference, MaskHand minimizes uncertainty in the 2D-to-3D mapping, enabling high-precision hand mesh recovery, even in occluded and ambiguous scenarios.


\begin{itemize}
    \item MaskHand is the first to leverage generative masked modeling for reconstructing robust 3D hand mesh. The main idea is to explicitly learn the  2D-to-3D probabilistic mapping and synthesize high-confidence, plausible 3D meshes by sampling from learned distribution.
    
    \item We design context-guided masked transformer that effectively fuses multiple contextual clues, including 2D pose, image features, and unmasked 3D pose tokens.
\item We propose differential masked training to learn 
      hand pose token distribution, conditioned on all contextual clues. This learned distribution facilitates confidence-guided sampling during inference, producing mesh reconstructions with low uncertainty and high precision.
    \item We show that MaskHand is a generic framework that unifies hand-mesh estimation and hand-mesh generation.
    \item We demonstrate through extensive experiments that MaskHand outperforms SOTA methods on standard datasets.
\end{itemize}
\vspace{-5pt}




%% file: sec/2_related.tex

\section{Related Work}

\subsection{Discriminative Methods}
Human hand recovery has been developed in recent years, with early approaches \cite{gao20223d, hoyet2012sleight, taylor2017articulated, wang2020rgb2hands, lightcap2021, zhao2012combining} leveraging optimization techniques to estimate hand poses based on 2D skeleton detections. 
Later, MANO \cite{MANO} introduced a differentiable parametric mesh model that capture hand shape and articulation, allowing the model to provide a plausible mesh with end-to-end estimation of model parameters directly from a single-view image.  Boukhayma et al. \cite{boukhayma20193d} presented the first fully learnable framework to directly predict the MANO hand model parameters \cite{MANO} from RGB images. Similarly, several subsequent methods have leveraged heatmaps \cite{zhang2019end} and iterative refinement techniques \cite{baek2019pushing} to ensure 2D alignment. Kulon et al. \cite{kulon2019single, kulon2020weakly} proposed a different regression approach that predicts 3D vertices instead of MANO pose parameters, achieving notable improvements over prior methods. Recent methods, such as METRO \cite{cho2022FastMETRO}, MeshGraphormer \cite{lin2021mesh}, HaMeR \cite{pavlakos2024reconstructing} achieve the SOTA performance  by modeling both vertex-vertex and vertex-joint interactions.  These existing methods, based on discriminative regression, learn a deterministic mapping from the input image to the output mesh. This deterministic approach struggles to capture the uncertainties and ambiguities caused by hand self-occlusions, interactions with objects, and extreme poses or camera angles, resulting in unrealistic hand mesh reconstructions.

\vspace{-5pt}

\subsection{Generative Methods}
Our MaskHand employs a generative method that learns a probabilistic mapping from the input image to the output mesh. It utilizes this learned distribution to synthesize high-confidence, plausible 3D hand meshes based on 2D visual contexts. HHMR \cite{li2024hhmr} is the only other generative hand mesh recovery method in the literature. Unlike HHMR \cite{li2024hhmr} that utilizes diffusion models, MaskHand is inspired by the success of masked image and language models for image and text generation tasks \cite{devlin2019bert,zhang2021ufc,chang2022maskgit,ding2022cogview2,chang2023muse}. This fundamental difference allows MaskHand to explicitly and quantitatively estimate confidence levels or prediction probabilities for all mesh reconstruction hypotheses, enabling confidence-guided hypothesis selection for accurate reconstruction, as shown in Fig. \ref{fig:MaskHand_conf_main}. In contrast, HHMR's denoising diffusion process synthesizes multiple mesh hypotheses without associating a confidence level with each hypothesis. Thus, it only reports the theoretically best mesh reconstruction by finding the hypothesis with minimal reconstruction errors under the assumption that ground-truth meshes are available.

%% file: sec/3_method.tex
\vspace{-7pt}
\section{Proposed Method: MaskHand}
\vspace{-7pt}

\partitle{Problem Formulation} Given a single-hand image \( I \), we aim to learn a mapping function \( f(I) = \{\theta, \beta, \pi \} \) that output the MANO \cite{MANO} model parameters from the input image. This mapping function encompasses three key components: the hand pose parameters \( \theta \in \mathbb{R}^{48} \), shape parameters \( \beta \in \mathbb{R}^{10} \), and camera parameters \( \pi \in \mathbb{R}^3 \), enabling comprehensive 3D hand reconstruction.

\begin{figure*}[ht]
    \centering
    \includegraphics[width=0.8\linewidth]{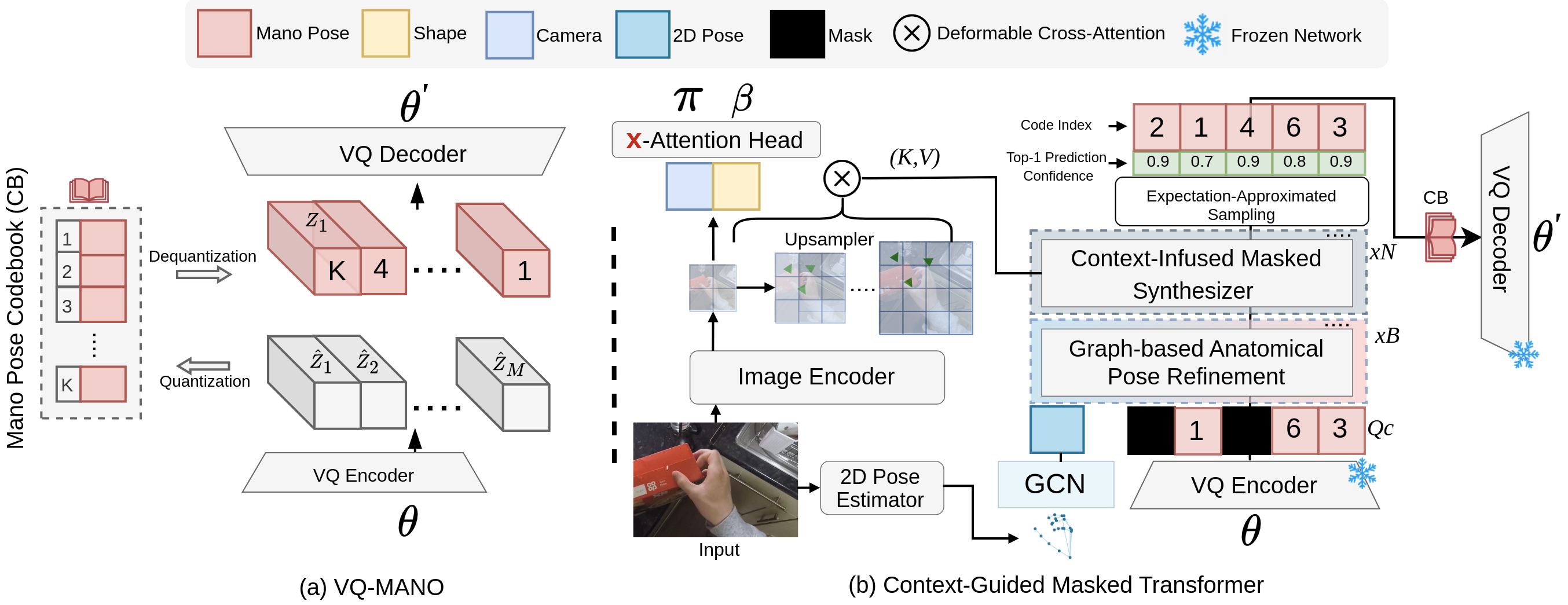}

\caption{MaskHand \textit{Training Phase}. MaskHand consists of two key components: (1) \textbf{VQ-MANO}, which encodes 3D hand poses into a sequence of discrete tokens within a latent space, and (2) a \textbf{Context-Guided Masked Transformer} that models the probabilistic distributions of these tokens, conditioned on the input image, 2D pose cues, and a partially masked token sequence.}    \label{fig:MaskHand_overview}
\end{figure*}

\partitle{Overview of Proposed Method} 
As shown in Figure \ref{fig:MaskHand_overview}, MaskHand consists of two primary modules: VQ-MANO and the Context-Guided Masked Transformer. Initially, VQ-MANO discretizes 3D MANO pose parameters (\(\theta\)) into discrete tokens, enabling generative masked modeling and uncertainty quantification. Next, the Context-Guided Masked Transformer leverages an encoder-decoder structure: the encoder extracts multi-scale image features, which, combined with 2D pose cues and unmasked tokens, are processed by the masked decoder. The decoder employs Graph-based Anatomical Pose Refinement to capture joint dependencies and a Context-Infused Masked Synthesizer to fuse visual features and token interactions, learning the probabilistic reconstruction via differential masked modeling. During inference, pose predictions are iteratively refined by selectively retaining high-confidence tokens and re-masking uncertain ones, progressively improving accuracy. Finally, the reconstruction is completed as the predicted pose (\( \theta \)) , shape  (\( \beta \)) and camera parameters (\( \pi \)) are feed into the MANO hand model. 


\begin{figure}[ht]
    \centering
    \includegraphics[width=0.9\linewidth]{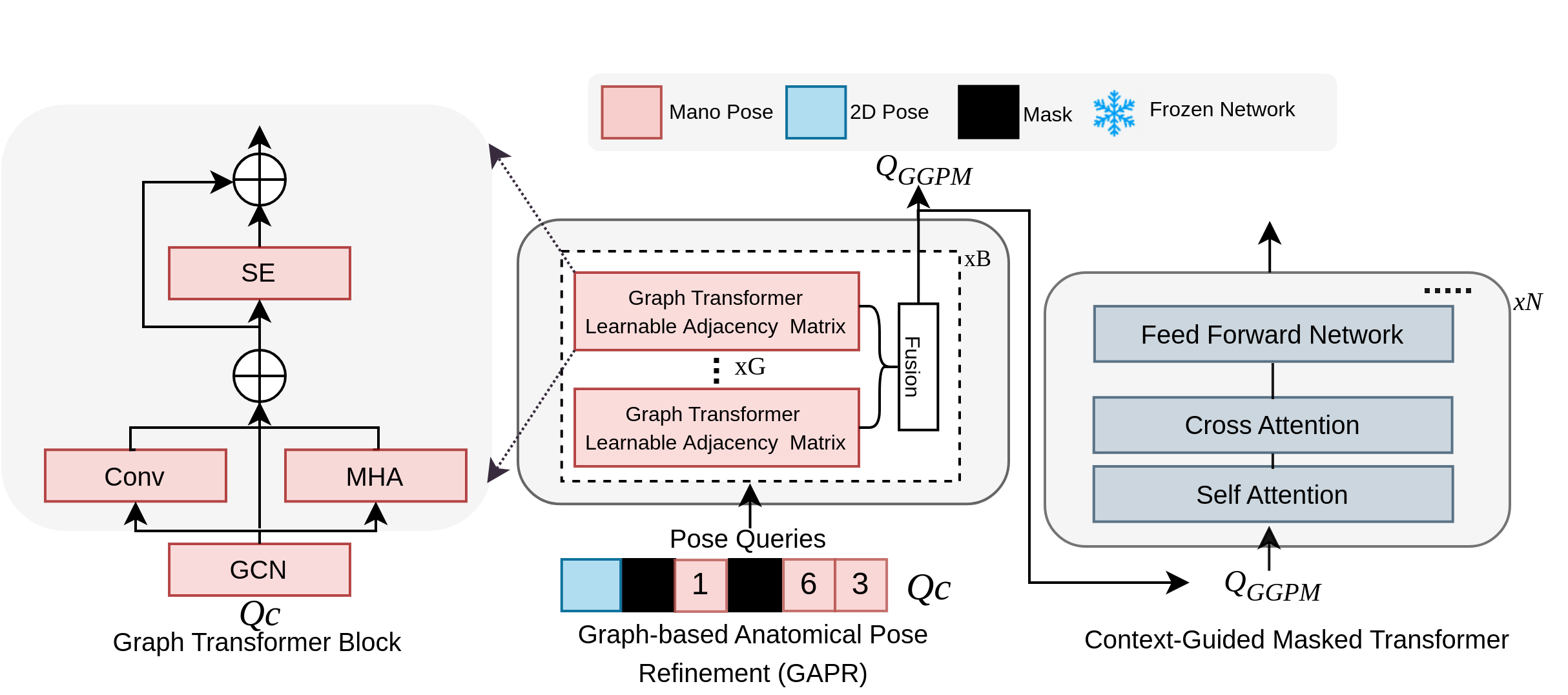}

\caption{Architecture of Graph-based Anatomical Pose Refinement (GAPR) and Context-Infused Masked Synthesizer, illustrating fusion of anatomical pose dependencies and contextual cues for precise mesh reconstruction.}

    \label{fig:GGPM_overview}
\end{figure}

\vspace{-5pt}

\subsection{Hand Model and VQ-MANO}

Our approach employs the MANO model \cite{MANO}, which maps pose \( \theta \in \mathbb{R}^{48} \) and shape \( \beta \in \mathbb{R}^{10} \) to a 3D hand mesh \( M \in \mathbb{R}^{778 \times 3} \) and joint locations \( X \in \mathbb{R}^{21 \times 3} \), representing both hand surface and pose.  

VQ-MANO is a MANO hand tokenizer that employs a 1D convolutional autoencoder to learn a discrete latent space for 3D pose parameters \( \theta \in \mathbb{R}^{48} \)  by quantizing the continuous pose embeddings into a learned codebook \( C \) with discrete code entries, as depicted in Figure~\ref{fig:MaskHand_overview}(a).  A key feature is its upsampling mechanism, expanding 16 MANO poses into 64 discrete pose tokens, which enhances spatial detail and improves mesh recovery accuracy (Supplementary Section C). To this end, we employ a Vector Quantized Variational Autoencoder (VQ-VAE) \cite{van2017neural} for pretraining the tokenizer. Specifically, we input the MANO pose parameters \( \theta \) into a convolutional encoder \( E \), which maps them to a latent embedding \( z \). Each embedding \( z_i \) is then quantized to its nearest codebook entry \( c_k \in C \) based on Euclidean distance, defined as
\[
\hat{z}_i = \arg\min_{c_k \in C} \| z_i - c_k \|_2.
\]

The total loss function of VQ-MANO is formulated as:
\[
\mathcal{L}_{\text{vq-mano}} = \lambda_{\text{re}} \mathcal{L}_{\text{recon}} + \lambda_{\text{E}} \| \operatorname{sg}[z] - c \|_2 + \lambda_{\alpha} \| z - \operatorname{sg}[c] \|_2,
\]
where \( \mathcal{L}_{\text{vq-mano}} \) consists of a MANO reconstruction loss \( \mathcal{L}_{\text{recon}} \), a latent embedding loss, and a commitment loss, weighted by hyperparameters \( \lambda_{\text{re}} \), \( \lambda_{\text{E}} \), and \( \lambda_{\alpha} \), respectively. Here, \( \operatorname{sg}[\cdot] \) denotes the stop-gradient operator, which prevents gradients from flowing through its argument during backpropagation. To further enhance reconstruction quality, we incorporate an additional L1 loss:
\[
\mathcal{L}_{\text{recon}} = \lambda_{\theta} \mathcal{L}_{\theta} + \lambda_{\text{V}} \mathcal{L}_{\text{V}} + \lambda_{\text{J}} \mathcal{L}_{\text{J}},
\]
which aims to minimize the discrepancies between the predicted and ground-truth MANO parameters, including the pose parameters \( \theta \), mesh vertices \( V \), and hand joints \( J \). The tokenizer is optimized using a straight-through gradient estimator to facilitate gradient propagation through the non-differentiable quantization step. Additionally, the codebook entries \( C \) are updated via exponential moving averages and periodic codebook resets, as described in \cite{esser2021taming,williams2020hierarchical}.

\vspace{-5pt}

\subsection{Context-Guided Masked Transformer} 

The context-guided masked transformer comprises two main components: the multi-scale image encoder and the masked graph transformer decoder.

\subsubsection{Multi-scale Image Encoder} Our encoder uses a vision transformer (ViT-H/16) \cite{alexey2020image} to extract image features \cite{pavlakos2024reconstructing}, processing 16x16 pixel patches into feature tokens. Following ViTDet \cite{li2022exploring}, we adopt a multi-scale feature approach by upsampling the initial feature map to produce feature maps at varying resolutions. This multi-scale representation is critical for handling complex articulations and self-occlusions in hand poses. High-resolution maps provide fine-grained joint details, while low-resolution maps capture global hand structure, balancing precision in joint positioning with overall anatomical coherence. Moreover, we utilize cross-attention with low-resolution feature maps in the x-Attention head to regress stable shape parameters (\( \beta \)) and camera orientation (\( \pi \)), making the process computationally efficient. This approach decouples shape estimation from pose modeling, preserving morphological stability and spatial alignment, and enhancing robustness and anatomical accuracy in 3D hand reconstruction.

\vspace{-5pt}

\subsubsection{Masked Graph Transformer Decoder}

The Masked Transformer Decoder is composed of two key components (Figure \ref{fig:MaskHand_overview} and \ref{fig:GGPM_overview}): Graph-based Anatomical Pose Refinement (GAPR) and the Context-Infused Masked Synthesizer Module. 

\partitle{Graph-based Anatomical Pose Refinement (GAPR)} Our decoder employs two blocks of a lightweight graph transformer that process pose tokens generated by VQ-MANO, enriched with 2D pose guidance, ensuring anatomically coherent hand reconstructions as shown in Figure \ref{fig:GGPM_overview}. To construct a hand-kinematic aware 2D pose token, we first process OpenPose 2D keypoints \cite{cao2017realtime} through a Graph Convolutional Network (GCN) with a fixed hand skeleton adjacency matrix, capturing explicit kinematic structures. Simultaneously, VQ-MANO pose tokens, which inherently encode implicit and non-linear joint dependencies, are extracted. These two representations are then fused and refined through two graph transformer block, where learnable adjacency matrices adaptively model complex joint relationships, ensuring robust spatial and structural consistency. To further enhance stability and anatomical accuracy, we integrate a transformer encoder with a Squeeze-and-Excitation (SE) block \cite{hu2018squeeze}, which emphasizes joint orientations and angles while ensuring spatial alignment through a 1x1 convolution layer. Within the transformer encoder, Multi-Head Attention (MHA) and pointwise convolution layers refine high-resolution dependencies between joints, producing a cohesive and anatomically aligned 3D pose representation, denoted as \( Q_{\text{GAPR}} \). The output \( Q_{\text{GAPR}} \) is formulated as:

\begin{equation}
Q' = \text{MHA}(\text{Norm}(Q_{\text{C}})) + \text{Conv}(\text{Norm}(Q_{\text{C}})) + Q_{\text{C}}
\end{equation}
\begin{equation}
Q_{\text{GAPR}} = \text{SE}(\text{Norm}(Q'))
\end{equation}
where \( Q_{\text{C}} \in \mathbb{R}^{K \times D} \) represents MANO pose tokens and 2D pose context queries, and \( Q_{\text{GAPR}} \) is the stabilized, anatomically consistent pose representation. These refined tokens \( Q_{\text{GAPR}} \) are then processed by the Context-Infused Masked Synthesizer to generate precise 3D hand meshes.



\partitle{Context-Infused Masked Synthesizer} We leverage a multi-layer transformer (as shown in Figure \ref{fig:GGPM_overview}) whose inputs are refined pose tokens \( Q_{\text{GAPR}} \) and cross-attends them with multi-scale feature maps generated by the image encoder. To enhance computational efficiency with high-resolution feature maps, a deformable cross-attention mechanism is employed \cite{Zhu2020DeformableDD}. This allows each pose token to focus on a selected set of sampling points around a learnable reference point, rather than the entire feature map. By concentrating attention on relevant areas, the model achieves a balance between computational efficiency and spatial precision, preserving essential information for accurate 3D hand modeling. The deformable cross-attention is defined as:
\[
\text{MCDA}(Q_{\text{GAPR}}, \hat{p}_y, \{x^l\}) = \sum_{l,k} A_{lyk} \cdot \mathbf{W} x^l (\hat{p}_y + \Delta p_{lyk})
\]
where \( Q_{\text{GAPR}} \) are refined manopose token queries, \( \hat{p}_y \) are learnable reference points, \( \Delta p_{lyk} \) are sampling offsets, \( \{x^l\} \) are multi-scale features, \( A_{lyk} \) are attention weights, and \( \mathbf{W} \) is a learnable weight matrix. With the inclusion of a \texttt{[MASK]} token in the masked transformer decoder, the module can predict masked pose tokens during training, while also facilitating token generation during inference. This approach uses the \texttt{[MASK]} token as a placeholder for pose predictions, enabling robust synthesis of occluded or unobserved hand parts for coherent 3D hand reconstruction.

\vspace{-5pt}

\subsection{Training: Differential Masked Modeling} 
\vspace{-3pt}

\partitle{Context-conditioned Masked Modeling} 
We employ masked modeling to train our model that learns the probabilistic distribution of 3D hand poses, conditioned on multiple contextual cues. Given a sequence of discrete pose tokens \( Y = [y_i]_{i=1}^L \) from the pose tokenizer, where \( L \) is the sequence length,  we randomly mask out a subset of $m$ tokens with \( m = \lceil \gamma(\tau) \cdot L \rceil \), where \( \gamma(\tau) \) is a cosine-based masking ratio function. Here, \( \tau \) is drawn from a uniform distribution \( U(0, 1) \), and we adopt the masking function \( \gamma(\tau) = \cos\left(\frac{\pi \tau}{2}\right) \), inspired by generative text-to-image modeling strategies \cite{chang2022maskgit}. 
Masked tokens are replaced with learnable \texttt{[MASK]} tokens, forming a corrupted sequence \( Y_{\overline{\mathbf{M}}} \) that the model must reconstruct. Each token \( y_i \) is predicted based on the probabilistic distribution \( p(y_i | Y_{\overline{\mathbf{M}}}, X_{2D}, X_{img}) \), conditioned on corrupted token sequence \( Y_{\overline{\mathbf{M}}} \), 2D pose embedding \( X_{2D} \), and image prompt $X_{img}$. This approach enables the model to explicitly account for the uncertainty inherent in mapping 2D observations to a coherent 3D hand mesh. The training objective is to minimize the negative log-likelihood of correctly predicting each pose token in the sequence, formulated as follows:

\[
\mathcal{L}_{\text{mask}} = -\mathbb{E}_{Y \in \mathcal{D}} \left[ \sum_{\forall i \in [1, L]} \log p(y_i | Y_{\overline{\mathbf{M}}}, X_{2D}, X_{img}) \right].
\]
\begin{figure}[ht]
    \centering
    \includegraphics[width=0.8\linewidth]{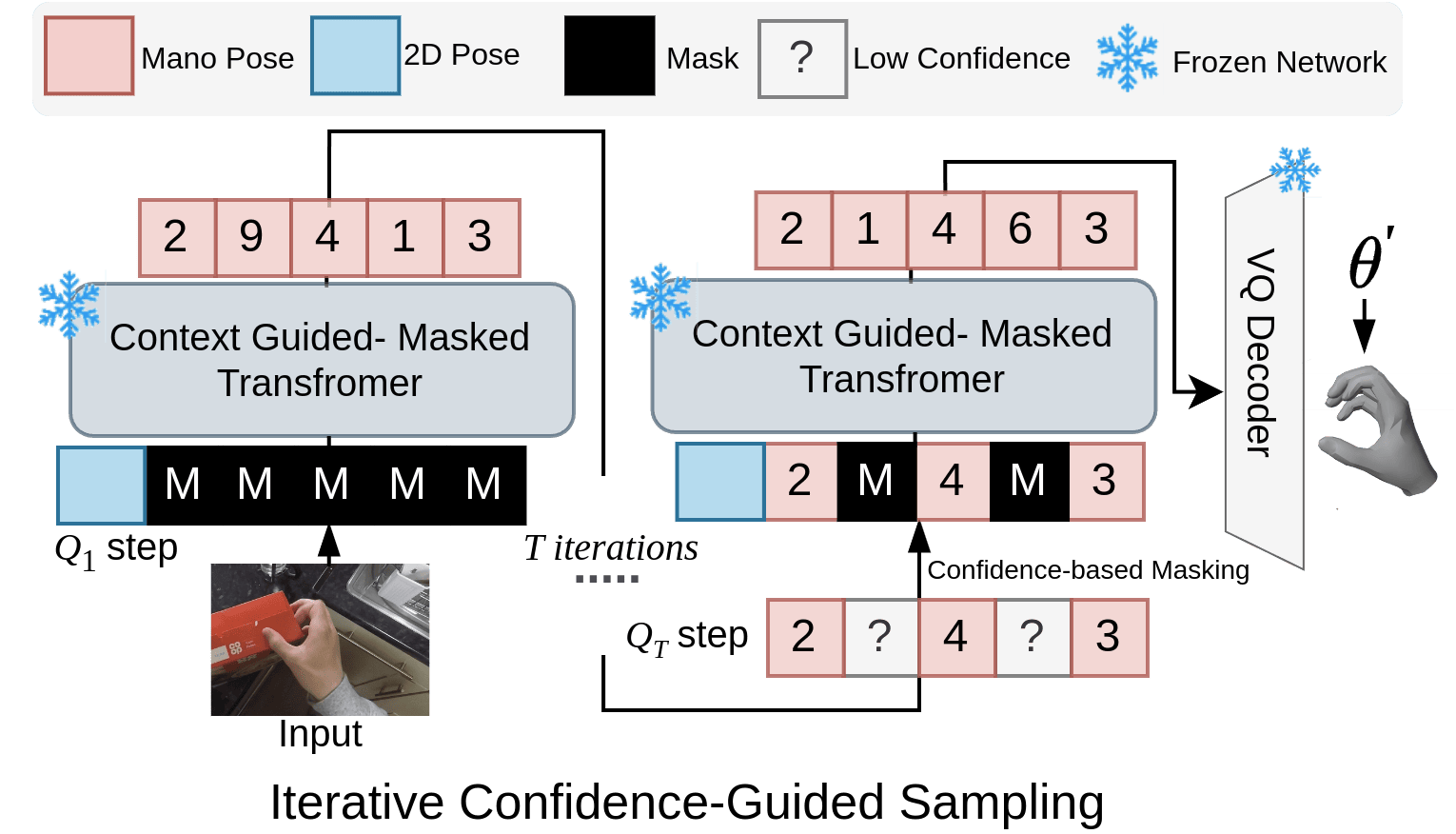}
\caption{MaskHand \textit{Inference Phase}: \textbf{Confidence-Guided Iterative Sampling} — a step-by-step refinement of pose selection by probabilistically sampling high-confidence tokens.}
    \label{fig:MaskHand_infer}
\end{figure}

\begin{figure*}[ht]
    \centering
    \includegraphics[width=0.80\linewidth]{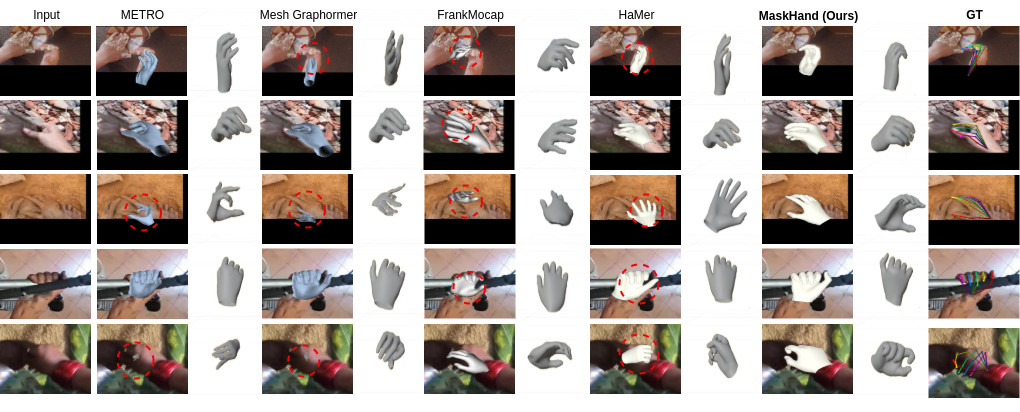}
\caption{Comparison of SOTA methods: MaskHand synthesizes unobserved parts for accurate 3D reconstructions in occluded scenarios.}
    \label{fig:MaskHand_sota}
\end{figure*}

\partitle{Expectation-Approximated Differential Sampling} The training objective \( \mathcal{L}_{\text{mask}} \) captures stochastic uncertainty in hand mesh reconstruction, enabling precise estimation of pose parameters \( \theta \) within a structured latent space. Recent studies \cite{pavlakos2024reconstructing} show that auxiliary 3D and 2D joint losses improve pose recovery by aligning predicted and ground-truth 2D/3D 
 joints. Integrating these losses requires transforming latent pose tokens into \( \theta \), a process involving non-differentiable probabilistic sampling.  To address this, instead of sampling the distribution \( p(y_i | Y_{\overline{\mathbf{M}}}, X_{2D}, X_{img}) \) to obtain the most probable token from codebook, we implement an expectation-based differential relaxation: instead of directly estimating discrete code indices, the model outputs logits \( L \) for each token. These logits undergo a softmax operation, producing the token distribution, which are multiplied by the pretrained codebook, resulting in the mean quantized feature representations \( \bar{z} = [ \bar{z}_1, \bar{z}_2, \dots, \bar{z}_M ] \):


\[
\bar{z} = \textnormal{softmax}(L_{M \times K}) \times \text{CB}_{K \times D} 
\]
where \( L \) represents the logits matrix, \( \text{CB} \) denotes the codebook, \( M \) is the token count, \( K \) specifies the codebook size, and \( D \) defines the dimensionality of each codebook entry. This mean token embeddings $\bar{z}$ are feed into the decoder to reconstruct the pose parameter $\theta'$. Combined with shape parameters $\beta$ and camera parameters $\pi$, this process reconstructs the 3D hand mesh, enabling the computation of both 3D joint loss and 2D projection loss. Since obtaining $\bar{z}$ is differential, the model can be trained end-to-end using the overall loss function .


\[\mathcal{L}_{\text{total}} = \mathcal{L}_{\text{mask}} + \mathcal{L}_{\text{MANO}} + \mathcal{L}_{\text{3D}} + \mathcal{L}_{\text{2D}} \] 
which combines the masked token prediction loss (\( \mathcal{L}_{\text{mask}} \)), 3D joint loss (\( \mathcal{L}_{\text{3D}} \)), 2D projection loss (\( \mathcal{L}_{\text{2D}} \)), and MANO parameter loss \( \mathcal{L}_{\text{MANO}} \).  Together, these terms collectively minimize discrepancies within MANO space, enabling precise 3D hand reconstruction.

\vspace{-7pt}

\subsection{Inference: Confidence-Guided Sampling} 
Our model leverages confidence-guided sampling to achieve precise and stable 3D hand pose predictions as shown in Figure \ref{fig:MaskHand_infer}. This process begins with a fully masked sequence \( Y_1 \) of length \( L \), with each token initialized as \texttt{[MASK]}. Over \( T \) decoding iterations, each iteration \( t \) applies stochastic sampling to predict masked tokens based on their distributions $p(y_i | Y_{\overline{\mathbf{M}}}, X_{2D}, X_{img})$.  Following each sampling step, tokens with the lowest prediction confidences are re-masked to be re-predicted in subsequent iterations. The number of tokens re-masked is determined by a masking schedule \( \lceil \gamma\left(\frac{t}{T}\right) \cdot L \rceil \), where \( \gamma \) is a decaying function of \( t \). This schedule dynamically adjusts masking intensity based on confidence, using a higher masking ratio in earlier iterations when prediction confidence is lower. Consequently, the model iteratively refines ambiguous regions, progressively improving prediction confidence as context builds. The masking ratio decreases with each step, stabilized by the cosine decay function \(\gamma\).

%% file: sec/4_experiment.tex
\vspace{-10pt}

\section{Experiments}
\vspace{-3pt}

\begin{table}[h!]
\centering
\vspace{-10pt}
\scalebox{0.65}{
\begin{tabular}{l|cc|cc|cc}
\toprule
\textbf{Method} & \makecell{PA-MPJPE \\ mm(↓)} & \makecell{PA-MPVPE \\ mm(↓)} & \makecell{F@5 \\ (↑)} & \makecell{F@15 \\ (↑)} & \makecell{AUC$_J$ \\ (↑)} & \makecell{AUC$_V$ \\ (↑)} \\
\midrule
S$^2$HAND \cite{chen2021model} & 11.5 & 11.1 & 0.448 & 0.932 & 0.769 & 0.778 \\
KPT-Transf. \cite{hampali2022keypoint} & 10.9 & - & - & - & 0.785 & - \\
ArtiBoost \cite{yang2022artiboost} & 10.8 & 10.4 & 0.507 & 0.946 & 0.785 & 0.792 \\
Yu et al. \cite{yu2022uv} & 10.8 & 10.4 & - & - & - & - \\
HandGCAT \cite{wang2023handgcat} & 9.3 & 9.1 & 0.552 & 0.956 & 0.814 & 0.818 \\
AMVUR \cite{jiang2023probabilistic} & 8.7 & 8.3 & 0.593 & 0.964 & 0.826 & 0.834 \\
HMP \cite{duran2024hmp} & 10.1 & - & - & - & - & - \\
SPMHand \cite{lu2024spmhand} & 8.8 & 8.6 & 0.574 & 0.962 & - & - \\
\midrule
\rowcolor{gray!15} \textbf{MaskHand} & \textbf{7.0} & \textbf{7.0} & \textbf{0.663} & \textbf{0.984} & \textbf{0.860} & \textbf{0.860} \\
\bottomrule
\end{tabular}
}
\caption{Zero-Shot 3D Mesh Reconstruction Evaluation on HO3Dv3 \cite{hampali2022keypoint} Benchmark: SOTA Comparison."-" indicates unreported metrics.}

\label{tab:ho3dv3_results}
\end{table}



\begin{table}[h!]
\centering

\vspace{-10pt}
\scalebox{0.67}{
\begin{tabular}{l|cc|cc}
\toprule
\textbf{Methods} & \makecell{PA-MPJPE \\ mm (↓)} & \makecell{PA-MPVPE \\ mm (↓)} & \makecell{F@5 \\ (↑)} & \makecell{F@15 \\ (↑)} \\
\midrule

MeshGraphormer$^\dagger$ \cite{lin2021mesh} & 5.9 & 6.0 & 0.764 & 0.986 \\
FastMETRO \cite{cho2022cross} & 6.5 & 7.1   & 0.687 & 0.983 \\
FastViT \cite{vasu2023fastvit} & 6.6 & 6.7   & 0.722 & 0.981 \\
AMVUR \cite{jiang2023probabilistic} & 6.2 & 6.1   & 0.767 & 0.987 \\
Deformer \cite{yoshiyasu2023deformable} & 6.2 & 6.4   & 0.743 & 0.984 \\
PointHMR \cite{kim2023sampling} & 6.1 & 6.6   & 0.720 & 0.984 \\
Zhou et al. \cite{zhou2024simple} &  5.7 & 6.0   & 0.772 & 0.986 \\
HaMeR \cite{pavlakos2024reconstructing} & 6.0 & 5.7   & 0.785 & 0.990 \\
HHMR$\mathsection$ \cite{li2024hhmr} & 5.8 & 5.8 & - & - \\
\midrule
\rowcolor{gray!15} \textbf{MaskHand$\ddagger$} & \textbf{5.5} & \textbf{5.4}   & \textbf{0.801} & \textbf{0.991} \\
\bottomrule
\end{tabular}
}
\caption{3D Mesh Reconstruction Accuracy on FreiHAND \cite{zimmermann2019FreiHAND}: SOTA Comparison. $^\dagger$ Denotes Test-Time Augmentation. PA-MPJPE/PA-MPVPE in mm. "-" indicates unreported metrics.}

\begin{flushleft}
\footnotesize{$\ddagger$ \textbf{MaskHand} produces confident mesh reconstruction, guided by the learned 2D-to-3D distribution \textbf{without} assuming GT meshes.}
    
    \footnotesize{$\mathsection$ HHMR reports best hypothesis mesh reconstruction with  minimum MPJPE and MPVPE, \textbf{with} the assumption of available GT meshes.} 
\end{flushleft}
\label{tab:FreiHAND_sota}

\end{table}

\begin{table}[h!]
\centering

\vspace{-10pt}
\scalebox{0.67}{
\begin{tabular}{l|cc|cc}
\toprule
\textbf{Method} & \makecell{PA-MPJPE \\ (↓)} & \makecell{PA-MPVPE \\ (↓)} & \makecell{MPJPE \\ (↓)} & \makecell{MPVPE \\ (↓)} \\
\midrule
METRO \cite{zhang2019end} & 7.0 & - & - & - \\
Spurr et al. \cite{spurr2020weakly} & 6.8 & - & - & - \\
Liu et al. \cite{Liu_2021_CVPR} & 6.6 & - & - & - \\
HandOccNet \cite{HandOccNet} & 5.8 & 5.5 & 14.0 & 13.1 \\
MobRecon \cite{chen2022mobrecon} & 6.4 & 5.6 & 14.2 & 13.1 \\
H2ONet \cite{xu2023h2onet} & 5.7 & 5.5 & 14.0 & 13.0 \\
Zhou et al. \cite{zhou2024simple} & 5.5 & 5.5 & 12.4 & 12.1 \\

\midrule
\rowcolor{gray!15} \textbf{Ours} & \textbf{5.0} & \textbf{4.9} & \textbf{11.7} & \textbf{11.2} \\
\bottomrule
\end{tabular}
}
\caption{3D Mesh Reconstruction Accuracy on DexYCB \cite{chao2021dexycb}: SOTA Comparison. "-" indicates unreported metrics.}

\label{tab:DexYCB_results}
\end{table}

\begin{table*}[h!]
\centering
\vspace{-10pt}
\scalebox{0.65}{
\begin{tabular}{ll|ccc|ccc|ccc}
\toprule
\textbf{Method} & \textbf{Venue} & \multicolumn{3}{c|}{\textbf{NewDays}} & \multicolumn{3}{c|}{\textbf{VISOR}} & \multicolumn{3}{c}{\textbf{Ego4D}} \\
\cmidrule{3-11}
& & \makecell{@0.05 (↑)} & \makecell{@0.1 (↑)} & \makecell{@0.15 (↑)} & \makecell{@0.05 (↑)} & \makecell{@0.1 (↑)} & \makecell{@0.15 (↑)} & \makecell{@0.05 (↑)} & \makecell{@0.1 (↑)} & \makecell{@0.15 (↑)} \\
\midrule

\midrule
\multicolumn{11}{c}{\textbf{Occluded Joints Evaluation}} \\
\midrule
FrankMocap \cite{rong2021frankmocap} & \textit{ICCVW 2021} & 9.2 & 28.0 & 46.9 & 11.0 & 33.0 & 55.0 & 8.4 & 26.9 & 45.1 \\
METRO \cite{lin2021end} & \textit{CVPR 2021} & 7.0 & 23.6 & 42.4 & 10.2 & 32.4 & 53.9 & 8.0 & 26.2 & 44.7 \\
MeshGraphormer \cite{lin2021mesh} & \textit{ICCV 2021} & 7.9 & 25.7 & 44.3 & 10.9 & 33.3 & 54.1 & 9.3 & 32.6 & 51.7 \\
HandOccNet (param) \cite{park2022handoccnet} & \textit{CVPR 2022} & 7.2 & 23.5 & 42.4 & 7.4 & 26.1 & 46.7 & 7.2 & 26.1 & 45.7 \\
HandOccNet (no param) \cite{park2022handoccnet} & \textit{CVPR 2022} & 9.8 & 31.2 & 50.8 & 9.9 & 33.7 & 55.4 & 9.6 & 31.1 & 52.7 \\
HaMeR \cite{pavlakos2024reconstructing} & \textit{CVPR 2024} & 27.2 & 60.8 & 78.9 & 25.9 & 60.8 & 80.7 & 23.0 & 56.9 & 76.3 \\
\midrule
\rowcolor{gray!15} \textbf{MaskHand} & \textbf{\textit{Ours}} & \textbf{29.4} & \textbf{64.1} & \textbf{80.3} & \textbf{31.4} & \textbf{64.2} & \textbf{83.6} & \textbf{29.4} & \textbf{65.3} & \textbf{80.1} \\
\midrule
\multicolumn{11}{c}{\textbf{Occluded Joints Evaluation with 80\% Masked Hand}} \\
\midrule
HaMeR \cite{pavlakos2024reconstructing} & \textit{CVPR 2024} & 8.7 & 28.4 & 48.9 & 7.1 & 24.4 & 41.7 & 7.1 & 23.4 &  40.2 \\
\midrule
\rowcolor{gray!15} \textbf{MaskHand} & \textbf{Ours} & \textbf{10.5} & \textbf{31.6} & \textbf{52.0} & \textbf{8.2} & \textbf{25.0} & \textbf{41.9} & \textbf{8.2} & \textbf{26.2} & \textbf{44.1} \\
\midrule
\multicolumn{11}{c}{\textbf{Occluded Joints Evaluation with 90\% Masked Hand}} \\
\midrule
HaMeR \cite{pavlakos2024reconstructing} & \textit{CVPR 2024} & 8.7 & 27.4 & 48.1 & 7.1 & 24.3 & 42.2 & 7.1 & 23.4 &  40.2 \\
\midrule
\rowcolor{gray!15} \textbf{MaskHand} & \textbf{Ours} & \textbf{10.4} & \textbf{31.0} & \textbf{51.1} & \textbf{9.1} & \textbf{27.3} & \textbf{44.3} & \textbf{8.0} & \textbf{25.1} & \textbf{43.2} \\

\bottomrule

\end{tabular}
}

\caption{Zero-shot evaluation on the HInt Benchmark \cite{pavlakos2024reconstructing} for occluded joint reconstruction using the PCK metric, comparing MaskHand to SOTA methods. MaskHand demonstrates superior robustness and resilience to missing hand structures under severe occlusions, maintaining strong performance even with 80\% and 90\% masked hand regions\textbf{None of the models were trained on the HInt dataset.}}

\label{tab:HInt_sota}
\end{table*}



\partitle{Training Datasets}To train the hand pose tokenizer, we employed a diverse set of datasets to capture a wide range of hand poses and interactions. This includes DexYCB \cite{chao2021dexycb}, InterHand2.6M \cite{moon2020interhand2}, MTC \cite{xiang2019monocular}, and RHD \cite{zimmermann2017learning}. For training MaskHand, we constructed a comprehensive dataset following a similar setup as \cite{pavlakos2024reconstructing} to ensure fair comparison. Specifically, for evaluations on HO3Dv3 \cite{hampali2022keypoint} and HInt \cite{pavlakos2024reconstructing}, MaskHand was trained on a diverse mix of datasets, including FreiHAND \cite{zimmermann2019FreiHAND}, HODv2 \cite{hampali2020honnotate}, MTC \cite{xiang2019monocular}, RHD \cite{zimmermann2017learning}, InterHand2.6M \cite{moon2020interhand2}, H2O3D \cite{hampali2020honnotate}, DexYCB \cite{chao2021dexycb}, COCO-Wholebody \cite{jin2020whole}, Halpe \cite{fang2022alphapose}, and MPII NZSL \cite{simon2017hand}. However, for FreiHAND and DexYCB, MaskHand was trained specifically on their \textbf{respective training datasets}.

\partitle{Evaluation Metrics}  We follow standard protocols \cite{pavlakos2024reconstructing} to assess 3D joint reconstruction using PA-MPJPE and \( \text{AUC}_{J} \). For 3D mesh evaluation, we use PA-MPVPE, \( \text{AUC}_{V} \), and F@5mm/F@15mm to measure vertex precision. Moreover, PCK evaluates 2D reconstruction by reprojecting 3D joints into 2D space. The computational efficiency is measured by Average Inference Time per Image (AITI) in seconds \cite{cao2017realtime}, where lower values indicate better performance.


\partitle{Reconstruction Evaluation} We evaluate MaskHand’s joint and mesh reconstruction on HO3Dv3 \cite{hampali2022keypoint}, DexYCB \cite{chao2021dexycb}, FreiHAND \cite{zimmermann2019FreiHAND}, and HInt \cite{pavlakos2024reconstructing} benchmarks.

\textit{\textbf{Zero-Shot Generalization Evaluation:}} The HO3Dv3 dataset, with 20K annotated images, serves as a benchmark for evaluating generalization in complex hand-object interactions. MaskHand is tested on HO3Dv3 \textbf{without prior training} to assess its capability in handling unseen data, occlusions, dataset biases, and diverse hand articulations in real-world scenarios.  Additionally, we evaluate MaskHand’s zero-shot performance on the HInt benchmark \cite{pavlakos2024reconstructing}, which presents challenges such as varied lighting, extreme viewpoints, and complex hand-object interactions. Using the PCK metric, MaskHand is tested on HInt-NewDays \cite{cheng2023towards}, HInt-EpicKitchensVISOR \cite{damen2018scaling}, and HInt-Ego4D \cite{grauman2022ego4d}, demonstrating its ability to operate in diverse and occlusion-heavy environments.


\textit{\textbf{Supervised Evaluation:}} We evaluated MaskHand on FreiHAND \cite{zimmermann2019FreiHAND}, which consists of 4K images captured in controlled environments, and DexYCB \cite{chao2021dexycb}, a large-scale dataset containing 78K images of hands interacting with objects. For a fair comparison, we trained MaskHand exclusively on FreiHAND, following the evaluation protocols of MeshGraphormer \cite{lin2021mesh} and HHMR \cite{li2024hhmr}. Similarly, aligning with recent work \cite{zhou2024simple}, we trained MaskHand on DexYCB and conducted evaluations under the same conditions, ensuring a direct and consistent comparison.

\vspace{-7pt}

\subsection{Comparison to State-of-the-art Approaches}
We evaluate MaskHand on HO3Dv3, FreiHAND, DexYCB, and HInt (Tables \ref{tab:ho3dv3_results}–\ref{tab:HInt_sota}), where it outperforms SOTA methods, demonstrating strong zero-shot generalization and in-the-wild robustness. Unlike prior approaches, MaskHand leverages confidence-aware masked modeling to effectively handle uncertainty, occlusions, and complex hand poses. In zero-shot evaluation on HO3Dv3 (Table \ref{tab:ho3dv3_results}), MaskHand reduces PA-MPJPE by 19.5\% and PA-MPVPE by 15.7\%, demonstrating superior dataset-agnostic reconstruction. On FreiHAND (Table \ref{tab:FreiHAND_sota}), it achieves 3.5\% lower PA-MPJPE and 5.3\% lower PA-MPVPE, with the highest F@5mm (0.801) and F@15mm (0.991), ensuring precise, high-fidelity hand recovery. In DexYCB (Table \ref{tab:DexYCB_results}), MaskHand reduces MPJPE by 9.1\% and MPVPE by 10.9\%, confirming its robustness in real-world hand-object interactions. In zero-shot evaluation on HInt (Table \ref{tab:HInt_sota}), which includes challenging egocentric, heavily occluded hand poses, MaskHand achieves state-of-the-art occluded joint reconstruction, improving PCK@0.05 by 8.1\% on HInt-NewDays, 21.2\% on HInt-VISOR, and 27.8\% on HInt-Ego4D.  Qualitatively, Figure~\ref{fig:MaskHand_sota} shows MaskHand’s superior synthesis of occluded regions compared to prior methods, while Figure~\ref{fig:mmhmr_maskhands_main} demonstrates its robustness under extreme masking. These results highlight MaskHand’s strong generalization and practical effectiveness in challenging real-world conditions.

%% file: sec/5_ablation_study.tex
\begin{figure}[ht] 
    \centering
    \includegraphics[width=0.75\linewidth]{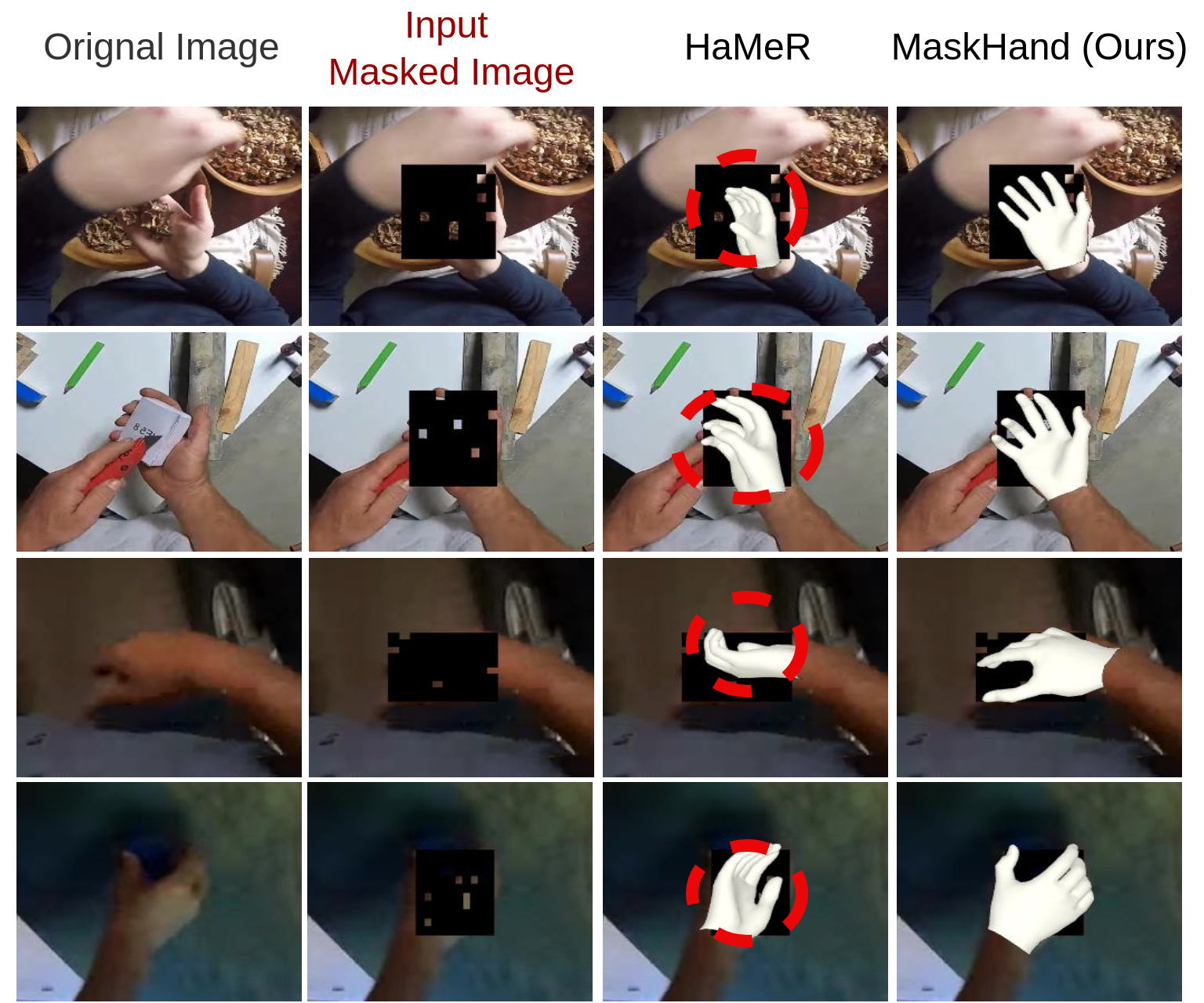}
    \caption{SOTA Comparison: Qualitative zero-shot evaluation on HInt Benchmark  \cite{pavlakos2024reconstructing} for heavily masked hand images. MaskHand reconstructs occluded hand poses, demonstrating robustness to severe occlusions and generalizability to unseen masked regions.}
    \label{fig:mmhmr_maskhands_main} 
\end{figure}
\vspace{-7pt}

\subsection{Ablation Study}


The key to MaskHand’s effectiveness lies in its generative masked modeling and uncertainty quantification, enabling robust reconstruction in occluded regions. Its iterative decoding strategy refines predictions, progressively enhancing accuracy in challenging scenarios. This ablation study explores iterative refinement and MaskHand’s versatility in text-conditioned 3D hand synthesis.

Further analyses, experimental details, and qualitative visualizations in the Supplementary Material explore key aspects, including the influence of \textit{\textbf{architectural choices}} (e.g., VQ-MANO tokenizer, proposed modules, multi-scale feature resolutions, deformable attention layers), \textit{\textbf{training strategies}} (masking ratios, regularization losses via differential sampling), and advanced \textit{\textbf{MaskHand capabilities}} such as unconditional mesh generation and robust reconstruction under severe real-world occlusions.

\partitle{Impact of Iterative Confidence-Guided Sampling} In confidence-guided sampling, we iteratively refines predictions by retaining high-confidence tokens and re-masking low-confidence ones, progressively enhancing stability and accuracy. As shown in Table \ref{tab:iter_decod}, increasing iterations from 1 to 5 reduces PA-MPVPE (7.2 $\rightarrow$ 7.0 on HO3Dv3, 5.6 $\rightarrow$ 5.4 on FreiHAND). These improvements result from confidence-guided masking of uncertain regions during decoding, enhancing reconstruction accuracy.



\begin{table}[h!]
\centering
\vspace{-7pt}
\scalebox{0.65}{
\begin{tabular}{l c c c c c}
\toprule
& \multicolumn{2}{c}{HO3Dv3} & \multicolumn{2}{c}{FreiHAND} & AITI (s) \\
\cmidrule(lr){2-3} \cmidrule(lr){4-5}
\textbf{\# of iter.} & PA-MPJPE  & PA-MPVPE & PA-MPJPE & PA-MPVPE & (↓)  \\
\midrule
1  & 7.2 & 7.2 & 5.6 & 5.6 & 0.04 \\
3  & 7.1 & 7.1 & 5.6 & 5.5 & 0.08 \\
5  & \textbf{7.0} & \textbf{7.0} & \textbf{5.5} & \textbf{5.4} & 0.12 \\
\bottomrule
\end{tabular}
}
\caption{Iterations in Iterative Confidence-Guided Sampling.  AITI is obtained on a single mid-grade GPU (NVIDIA RTX A5000).}

\label{tab:iter_decod}
\end{table}

\vspace{-7pt}

\subsection{Application: Text-to-Mesh Generation}
\label{sec:gen_t2m}
\begin{figure}[ht] 
    \centering
    \includegraphics[width=0.8\linewidth]{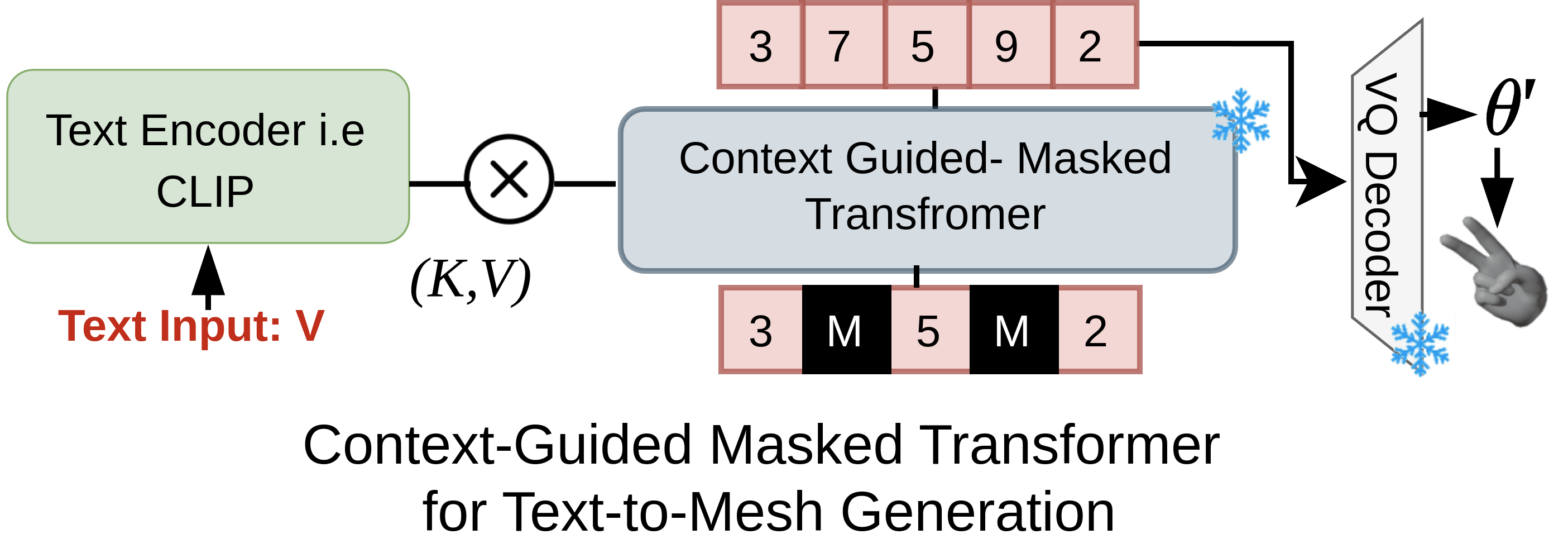}
   \caption{
    Repurposed MaskHand for Text-to-Mesh Generation
}    
    \label{fig:openclip}
\end{figure}

MaskHand’s modular design extends beyond hand mesh reconstruction to text-conditioned 3D hand synthesis. By replacing the image encoder with a CLIP-based text encoder \cite{radford2021learning}, MaskHand can generate 3D hand meshes from text inputs, as shown in Figure~\ref{fig:openclip}. Using the ASL dataset \cite{kaggle_asl_alphabet}, we trained MaskHand with pseudo-ground-truth annotations generated by MaskHand. During testing, a top-5\% probabilistic sampling strategy ensured diverse yet plausible mesh outputs, evaluated using Hausdorff Distance, Chamfer Distance, and PA-MPVPE. Results (Table~\ref{tab:text_to_mesh_metrics}) demonstrate high fidelity and consistency in text-driven reconstruction, with qualitative examples shown in Figure~\ref{fig:ASL_alphabet}. More details are provided in Supplementary Section B.

\begin{table}[h!]
\centering

\vspace{5pt}
\scalebox{0.6}{
\begin{tabular}{lcc}
\toprule
\textbf{Metric} & \textbf{Mean} & \textbf{Standard Deviation} \\
\midrule
Hausdorff Distance & 0.0221 & 0.0073 \\
Chamfer Distance & $9.73 \times 10^{-5}$ & $5.47 \times 10^{-5}$ \\
PA-MPVPE (mm) & 12.2 & 3.1 \\
\bottomrule
\end{tabular}
}
\caption{Quantitative evaluation of text-to-mesh generation using MaskHand's Masked Synthesizer, reporting mean and standard deviation across all generated meshes per text prompt.}

\label{tab:text_to_mesh_metrics}
\end{table}

\begin{figure}[ht] 
    \centering
    \includegraphics[width=0.8\linewidth]{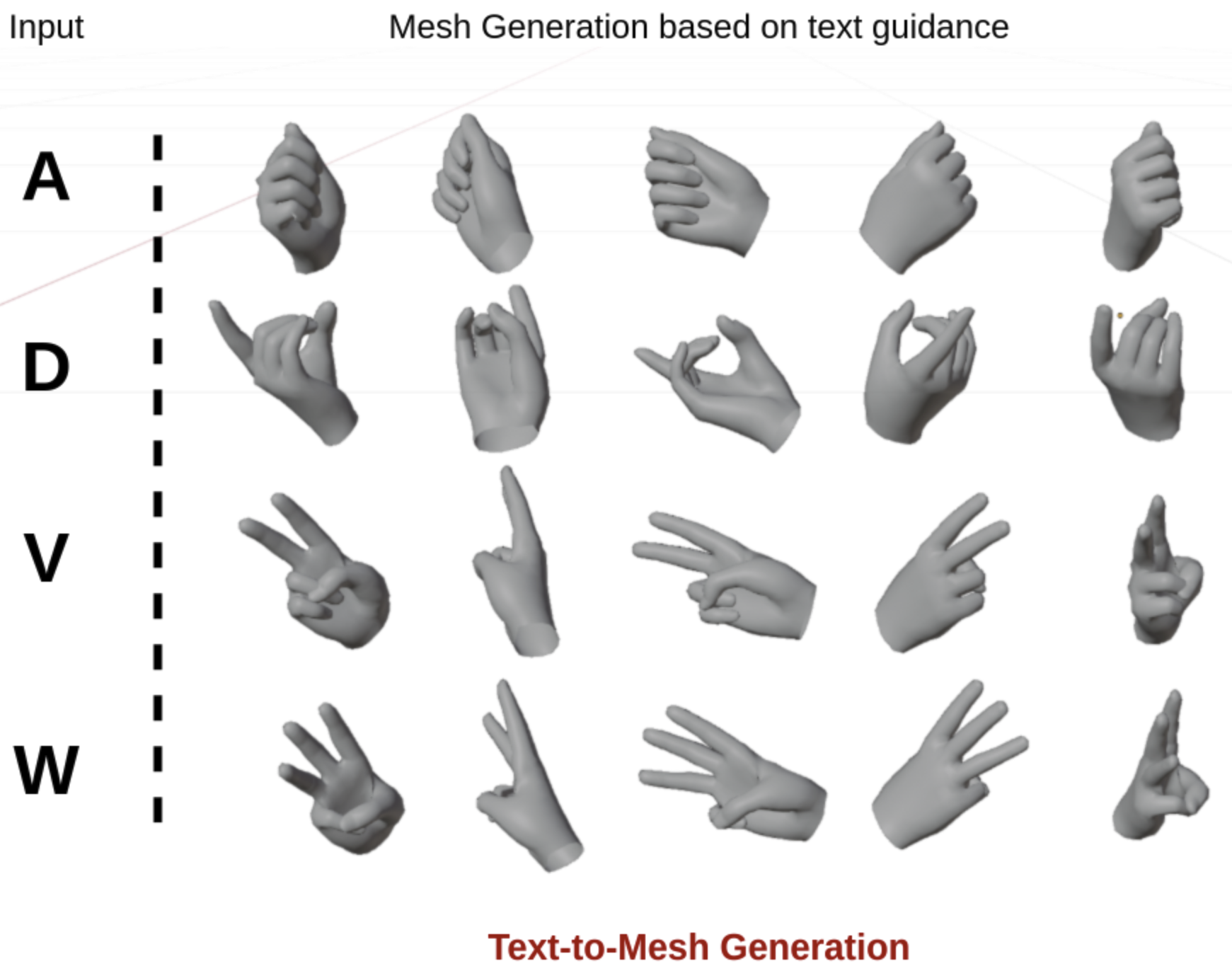}
    \caption{Text-to-mesh generation results on the ASL dataset, demonstrating MaskHand’s Masked Synthesizer's ability to generate structurally accurate and diverse meshes from text inputs.}
    \label{fig:ASL_alphabet}
\end{figure}

%% file: sec/6_conclusion.tex
\vspace{-10pt}

\section{Conclusion}
We introduced MaskHand, a generative masked model for accurate and robust 3D hand mesh reconstruction, particularly in occluded and ambiguous regions. By explicitly learning and sampling from the probabilistic 2D-to-3D mapping, MaskHand effectively handles self-occlusions, depth ambiguities, and missing hand parts. At its core, VQ-MANO encodes 3D hand articulations as discrete pose tokens, enabling structured latent representations. The Context-Guided Masked Transformer models joint token dependencies by incorporating image context, 2D pose cues, and masked token distributions. Through confidence-guided iterative sampling, MaskHand progressively refines reconstructions, selectively updating uncertain regions to produce anatomically coherent and high-precision hand meshes. Extensive evaluations demonstrate that MaskHand outperforms state-of-the-art methods, particularly in challenging occlusion-heavy scenarios. This work sets a new benchmark for 3D hand modeling, with broad applications in human-computer interaction, AR, and VR.



%% file: sec/X_suppl.tex
\clearpage
\onecolumn




\section{Supplementary Material}
\appendix

\section{Overview}

The supplementary material is organized into the following sections:

\begin{itemize}
    \item  Section \ref{sec:implementation}: Implementation Details
    \item  Section \ref{sec:vqmano}: Ablation for VQ-MANO Pose Tokenizer 
    \item Section \ref{sec:hint_sota}: In-the-Wild Reconstruction Evaluation
    \item Section \ref{sec:mask_hands_supp}: Occluded and Masked Hands Reconstruction 
    \item Section \ref{sec:comp_anal_eff}: Effectiveness of Proposed MaskHand Components
    \item  Section \ref{sec:pose2d_est}: Impact of 2D Pose Context
    \item  Section \ref{sec:uncond_gen}: Confidence-Aware Unconditional Mesh Generation 
    \item  Section \ref{sec:mask_ratio}: Masking Ratio during Training
    \item  Section \ref{sec:diffsampling}: Effectiveness of Expectation-Approximated Differential Sampling
    \item  Section \ref{sec:conf_guided}: Confidence-Guided Masking
    \item  Section \ref{sec:mano_MaskHand}: Impact of VQ-MANO Tokenizer on MaskHand
    \item  Section \ref{sec:feats_res}: Impact of Multi-Scale Features
    \item  Section \ref{sec:deformable}: Deformable Cross-Attention Layers in  MaskHand
    \item Section \ref{sec:qualitative}: Qualitative Results in the Wild

\end{itemize}
Project website can be founds at \url{https://m-usamasaleem.github.io/publication/MaskHand/MaskHand.html}.

\section{Implementation Details}
\label{sec:implementation}

The implementation of MaskHand, developed using PyTorch, comprises two essential training phases: the VQ-MANO tokenizer and the context-guided masked transformer. These phases are meticulously designed to ensure accurate 3D hand mesh reconstruction while balancing computational efficiency and model robustness.

\partitle{VQ-MANO} In the first phase, the VQ-MANO module is trained to learn discrete latent representations of hand poses. The pose parameters, \( \theta \in \mathbb{R}^{16 \times 3} \), encapsulate the global orientation (\( \theta_1 \in \mathbb{R}^3 \)) and local rotations (\( [\theta_2, \ldots, \theta_{16}] \in \mathbb{R}^{16 \times 3} \)) of hand joints. The architecture of the tokenizer employs ResBlocks \cite{he2016deep} and 1D convolutional layers for the encoder and decoder, with a single quantization layer mapping continuous embeddings into a discrete latent space. To train the hand pose tokenizer, we utilized a range of datasets capturing diverse hand poses, interactions, and settings. Specifically, we leveraged DexYCB \cite{chao2021dexycb}, InterHand2.6M \cite{moon2020interhand2}, MTC \cite{xiang2019monocular}, and RHD \cite{zimmermann2017learning}. These datasets collectively provide a rich spectrum of annotated data, enabling the model to generalize effectively across various real-world scenarios. The training process spans 400K iterations and uses the Adam optimizer with a batch size of 512 and a learning rate of \( 1 \times 10^{-4} \). The loss function combines reconstruction and regularization objectives, weighted as \( \lambda_{\text{recon}} = 1.0 \), \( \lambda_{\text{E}} = 0.02 \), \( \lambda_{\theta} = 1.0 \), \( \lambda_{\text{V}} = 0.5 \), and \( \lambda_{\text{J}} = 0.3 \). The final pose tokenizer is trained on DexYCB, InterHand2.6M, MTC, and RHD datasets, resulting in a model with 64 tokens and a codebook size of \( 2048 \times 256 \).

\partitle{Context-Guided Masked Transformer}
The second phase involves training the context-guided masked transformer, with the pose tokenizer frozen to leverage its pre-trained pose priors. This phase is dedicated to synthesizing pose tokens conditioned on input images and refining the 3D mesh reconstruction. Multi-resolution feature maps at \( 1\times \) and  \( 4\times \) scales are used to capture both global and local contextual details, allowing the model to handle complex hand articulations and occlusions. The overall architecture of the system, including the Graph-based Anatomical Pose Refinement (GAPR) and Context-Infused Masked Synthesizer.  The GAPR consists of two blocks (x$B$) of graph transformers to effectively model joint dependencies and ensure anatomical consistency. Meanwhile, the Context-Infused Masked Synthesizer employs four transformer layers ($xN$) to integrate multi-scale image features and refine pose token predictions through deformable cross-attention and token dependencies. The default number of iterations in Confidence-Guided Sampling is 5, which we use for the ablation study. The overall loss function integrates multiple objectives to guide the model toward robust reconstructions:
\[
\mathcal{L}_{\text{total}} = \mathcal{L}_{\text{mask}} + \mathcal{L}_{\text{MANO}} + \mathcal{L}_{\text{3D}} + \mathcal{L}_{\text{2D}},
\]
where \( \mathcal{L}_{\text{mask}} \) minimizes errors in masked token predictions, \( \mathcal{L}_{\text{MANO}} \) ensures consistency in MANO shape (\( \beta \)) and pose (\( \theta \)) parameters, \( \mathcal{L}_{\text{3D}} \) aligns the predicted and ground-truth 3D joint positions, and \( \mathcal{L}_{\text{2D}} \) preserves accurate 2D joint projections. The loss weights are configured as \( \lambda_{\text{mask}} = 1.0 \), \( \lambda_{\text{MANO}} = 1.5 \times 10^{-3} \) (with \( \lambda_{\theta} = 1 \times 10^{-3} \) for pose and \( \lambda_{\beta} = 5 \times 10^{-4} \) for shape), \( \lambda_{\text{3D}} = 5 \times 10^{-2} \), and \( \lambda_{\text{2D}} = 1 \times 10^{-2} \). This phase is trained for 200K iterations using the Adam optimizer on NVIDIA RTX A6000 GPUs with a batch size of 48 and a learning rate of \( 1 \times 10^{-5} \).

\partitle{Generalization to Text-to-Mesh Generation Details}
The MaskHand model is designed to be modular and adaptable, extending beyond image-conditioned tasks to support text-to-mesh generation. To achieve this, we replaced image-based conditioning with text guidance, enabling MaskHand’s Masked Synthesizer to generate diverse 3D meshes directly from textual input. Additionally, we explored the model’s capability to synthesize meshes without 2D pose guidance, making it rely solely on text prompts for generation. For training, we used the American Sign Language (ASL) dataset, a widely recognized resource for hand-based sign language recognition in English-speaking regions such as the United States and Canada. The dataset consists of 26 one-handed gestures representing the alphabet, making it suitable for text-to-mesh experiments. Specifically, we used the ASL alphabet dataset from Kaggle \cite{kaggle_asl_alphabet} for training. Since the ASL dataset lacks 3D annotations (e.g., MANO parameters), we leveraged MaskHand to generate pseudo-ground-truth (p-GT) annotations, which were then used to train the text-guided version of the model. To integrate textual information, we extracted CLIP \cite{radford2021learning} embeddings from ASL labels, enabling seamless text-based conditioning within the generative pipeline. During testing, we applied a top-5\% probabilistic sampling strategy, allowing the model to generate multiple plausible meshes per text input while ensuring diversity and consistency in synthesis.

\subsection{Data Augmentation}

In the initial training phase, the VQ-MANO  module leverages prior knowledge of valid hand poses, serving as a critical foundation for the robust performance of the overall MaskHand pipeline. To deepen the model's understanding of pose parameters, hand poses are systematically rotated across diverse angles, enabling it to effectively learn under varying orientations. In the subsequent training phase, the robustness of  MaskHand  is further enhanced through an extensive augmentation strategy applied to both input images and hand poses. These augmentations—such as scaling, rotations, random horizontal flips, and color jittering—introduce significant variability into the training data. By simulating real-world challenges like occlusions and incomplete pose information, these transformations prepare the model for complex, unpredictable scenarios. This comprehensive approach to data augmentation is a cornerstone of the training process, significantly improving the model’s ability to generalize and produce reliable, precise 3D hand mesh reconstructions across a wide range of conditions.

\subsection{Camera Model} 
In the MaskHand pipeline, a simplified perspective camera model is employed to project 3D joints onto 2D coordinates, striking a balance between computational efficiency and accuracy. The camera parameters, collectively represented by \( \Pi \), include a fixed focal length, an intrinsic matrix \( K \in \mathbb{R}^{3 \times 3} \), and a translation vector \( T \in \mathbb{R}^3 \). To streamline computations, the rotation matrix \( R \) is replaced with the identity matrix \( I_3 \), further simplifying the model. The projection of 3D joints \( J_{\text{3D}} \) onto 2D coordinates \( J_{\text{2D}} \) is described as \( J_{\text{2D}} = \Pi(J_{\text{3D}}) \), where the operation encapsulates both the intrinsic parameters and the translation vector. This modeling approach reduces the parameter space, enabling computational efficiency while maintaining the accuracy required for robust 3D hand mesh reconstruction. By focusing on the most critical components, the model minimizes complexity without compromising performance.

\section{Ablation for VQ-MANO}
\label{sec:vqmano}

Tables \ref{tab:tokenizer_tokens} and \ref{tab:tokenizer_cb} summarize an ablation study on the Freihand \cite{zimmermann2019FreiHAND} dataset, focusing on two key parameters: the number of pose tokens and the codebook size. Table \ref{tab:tokenizer_tokens} shows that increasing the number of pose tokens, while fixing the codebook size at \(2048 \times 256\), improves performance significantly, reducing PA-MPJPE from 1.01 mm to 0.41 mm and PA-MPVPE from 0.97 mm to 0.41 mm as tokens increase from 16 to 128. Table \ref{tab:tokenizer_cb} highlights the effect of increasing the codebook size with a fixed token count of 64, showing a reduction in PA-MPJPE from 0.66 mm to 0.43 mm and PA-MPVPE from 0.65 mm to 0.44 mm as the size grows from \(1024 \times 256\) to \(4096 \times 256\). Notably, the codebook size has a stronger impact on performance than the number of pose tokens. The final configuration, with a codebook size of \(2048 \times 256\) and 64 tokens, balances efficiency and accuracy, achieving PA-MPJPE of 0.47 mm and PA-MPVPE of 0.44 mm. These results emphasize the importance of jointly optimizing these parameters for effective hand pose tokenization.


\begin{table}[h!]
\caption{Impact of Number of Pose Tokens (Codebook = 2048 $\times$ 256) on VQ-MANO on Freihand dataset}
\vspace{-7pt}
\centering
\scalebox{1}{
\begin{tabular}{c c c c c}
\toprule
 & \multicolumn{4}{c}{Number of Pose Tokens}  \\ \cmidrule(lr){2-5} 
\textbf{Metric} & \textbf{16} & \textbf{32} & \textbf{64} & \textbf{128} \\
\midrule
PA-MPJPE (mm) & 1.01 & 0.59 & 0.47 & 0.41\\
PA-MPVPE (mm) & 0.97 & 0.57 & 0.44 & 0.41 \\
\bottomrule
\end{tabular}
}
\centering
\centering
\label{tab:tokenizer_tokens}
\end{table}

\begin{table}[h!]
\caption{Impact of Number of Codebook Size (Pose Tokens = 64) on VQ-MANO on Freihand dataset}

\vspace{-7pt}
\centering
\scalebox{1}{
\begin{tabular}{c c c c c}
\toprule
 & \multicolumn{4}{c}{Number of Codebook Size}  \\ \cmidrule(lr){2-5} 
\textbf{Metric} & \textbf{1024 × 256} & \textbf{2048 × 128} & \textbf{2048 × 256} & \textbf{4096 × 256} \\
\midrule
PA-MPJPE (mm) & 0.66 & 0.56 & 0.47 & 0.43\\
PA-MPVPE (mm) & 0.65 & 0.58 & 0.44 & 0.44 \\
\bottomrule
\end{tabular}
}
\centering
\centering
\label{tab:tokenizer_cb}
\end{table}

\begin{figure}[ht] 
    \centering
    \includegraphics[width=1\linewidth]{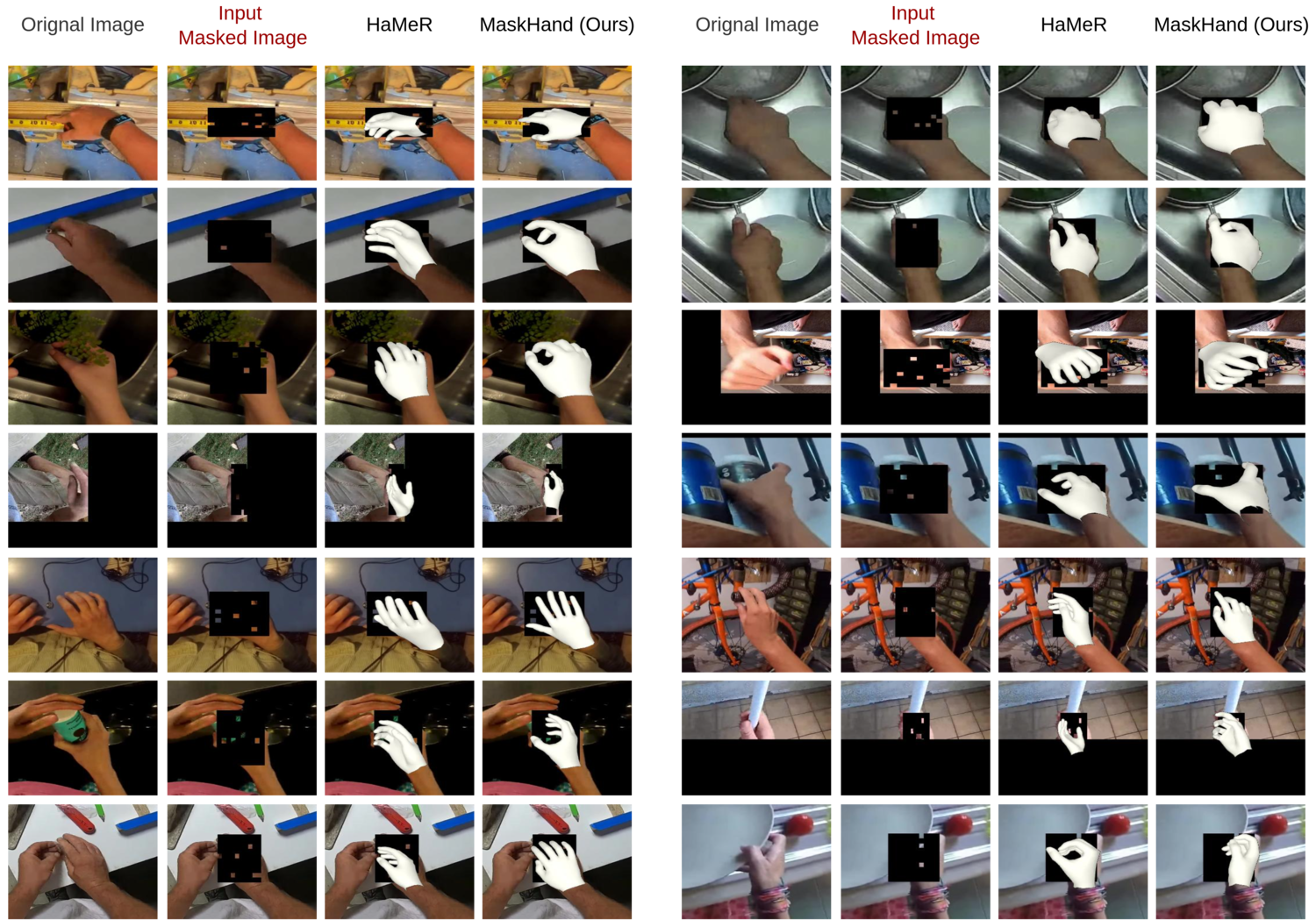}
    \caption{SOTA Comparison: Qualitative zero-shot evaluation on HInt Benchmark  \cite{pavlakos2024reconstructing} for heavily masked hand images. MaskHand reconstructs occluded hand poses, demonstrating robustness to severe occlusions and generalizability to unseen masked regions.}
    \label{fig:mmhmr_maskhands} 
\end{figure}

\section{In-the-Wild Reconstruction Evaluation}
\label{sec:hint_sota}
Table~\ref{tab:HInt_sota_supp} presents a zero-shot evaluation comparing MaskHand with recent state-of-the-art methods on the challenging HInt benchmark using the PCK metric. MaskHand consistently achieves the best results across all subsets—NewDays, VISOR, and Ego4D—and evaluation criteria (All Joints and Visible Joints). Notably, MaskHand significantly surpasses HaMeR (previous SOTA), demonstrating improvements of up to 7.5\% at the strictest threshold (PCK@0.05) on Ego4D (46.4\% vs. HaMeR's 38.9\%) and 3.1\% on VISOR (46.1\% vs. 43.0\%) for All Joints. Similar trends appear with Visible Joints, where MaskHand improves by 5.6\% on VISOR (62.1\% vs. HaMeR's 56.6\%) and 7.3\% on Ego4D (59.3\% vs. 52.0\%). These substantial gains underscore MaskHand's superior accuracy and robustness in reconstructing hands under challenging, real-world occlusions. While the model demonstrates impressive performance, the results also highlight areas for future improvements, particularly in severely occluded regions, where further advancements could provide additional gains in accuracy.

\begin{table}[h!]
\centering
\caption{Zero-Shot In-the-Wild Robustness Evaluation on the HInt Benchmark \cite{pavlakos2024reconstructing} using the PCK Metric: Comparison with SOTA Methods. \textbf{None of the models were trained on or have previously seen the HInt dataset.} }
\vspace{-10pt}
\scalebox{0.75}{
\begin{tabular}{ll|ccc|ccc|ccc}
\toprule
\textbf{Method} & \textbf{Venue} & \multicolumn{3}{c|}{\textbf{NewDays}} & \multicolumn{3}{c|}{\textbf{VISOR}} & \multicolumn{3}{c}{\textbf{Ego4D}} \\
\cmidrule{3-11}
& & \makecell{@0.05 (↑)} & \makecell{@0.1 (↑)} & \makecell{@0.15 (↑)} & \makecell{@0.05 (↑)} & \makecell{@0.1 (↑)} & \makecell{@0.15 (↑)} & \makecell{@0.05 (↑)} & \makecell{@0.1 (↑)} & \makecell{@0.15 (↑)} \\
\midrule
\multicolumn{11}{c}{\textbf{All Joints}} \\
\midrule
FrankMocap \cite{rong2021frankmocap} & \textit{ICCVW 2021} & 16.1 & 41.4 & 60.2 & 16.8 & 45.6 & 66.2 & 13.1 & 36.9 & 55.8 \\
METRO \cite{lin2021end} & \textit{CVPR 2021} & 14.7 & 38.8 & 57.3 & 16.8 & 45.4 & 65.7 & 13.2 & 35.7 & 54.3 \\
MeshGraphormer \cite{lin2021mesh} & \textit{ICCV 2021} & 16.8 & 42.0 & 59.7 & 19.1 & 48.5 & 67.4 & 14.6 & 38.2 & 56.0 \\
HandOccNet (param) \cite{park2022handoccnet} & \textit{CVPR 2022} & 9.1 & 28.4 & 47.8 & 8.1 & 27.7 & 49.3 & 7.7 & 26.5 & 47.7 \\
HandOccNet (no param) \cite{park2022handoccnet} & \textit{CVPR 2022} & 13.7 & 39.1 & 59.3 & 12.4 & 38.7 & 61.8 & 10.9 & 35.1 & 58.9 \\
HaMeR \cite{pavlakos2024reconstructing} & \textit{CVPR 2024} & 48.0 & 78.0 & 88.8 & 43.0 & 76.9 & 89.3 & 38.9 & 71.3 & 84.4 \\
\midrule

\rowcolor{gray!15} \textbf{MaskHand} & \textbf{Ours} & \textbf{48.7} & \textbf{79.2} & \textbf{90.0} & \textbf{46.1} & \textbf{81.4} & \textbf{92.1} & \textbf{46.4} & \textbf{77.5} & \textbf{90.1} \\
\midrule
\multicolumn{11}{c}{\textbf{Visible Joints}} \\
\midrule
FrankMocap \cite{rong2021frankmocap} & \textit{ICCVW 2021} & 20.1 & 49.2 & 67.6 & 20.4 & 52.3 & 71.6 & 16.3 & 43.2 & 62.0 \\
METRO \cite{lin2021end} & \textit{CVPR 2021} & 19.2 & 47.6 & 66.0 & 19.7 & 51.9 & 72.0 & 15.8 & 41.7 & 60.3 \\
MeshGraphormer \cite{lin2021mesh} & \textit{ICCV 2021} & 22.3 & 51.6 & 68.8 & 23.6 & 56.4 & 74.7 & 18.4 & 45.6 & 63.2 \\
HandOccNet (param) \cite{park2022handoccnet} & \textit{CVPR 2022} & 10.2 & 31.4 & 51.2 & 8.5 & 27.9 & 49.8 & 7.3 & 26.1 & 48.0 \\
HandOccNet (no param) \cite{park2022handoccnet} & \textit{CVPR 2022} & 15.7 & 43.4 & 64.0 & 13.1 & 39.9 & 63.2 & 11.2 & 36.2 & 56.0 \\
HaMeR \cite{pavlakos2024reconstructing} & \textit{CVPR 2024} & 60.8 & 87.9 & 94.4 & 56.6 & 88.0 & 94.7 & 52.0 & 83.2 & 91.3 \\
\midrule

\rowcolor{gray!15} \textbf{MaskHand} & \textbf{Ours} & \textbf{61.0} & \textbf{87.1} & \textbf{94.8} & \textbf{62.1} & \textbf{90.2} & \textbf{95.0} & \textbf{59.3} & \textbf{88.3} & \textbf{94.4} \\
\midrule

\end{tabular}
}
\label{tab:HInt_sota_supp}
\end{table}

\section{Occluded and Masked Hands Reconstruction}
\label{sec:mask_hands_supp}
Figure~\ref{fig:mmhmr_maskhands} provides a qualitative, zero-shot comparison between MaskHand and the state-of-the-art HaMeR method on severely masked images (with hand regions masked around 90\%). Despite significant occlusion, MaskHand consistently generates more plausible and anatomically accurate hand reconstructions than HaMeR. Specifically, MaskHand effectively synthesizes occluded regions, demonstrating robust generalization to previously unseen masked areas and preserving natural hand poses and orientations. In contrast, HaMeR exhibits noticeable reconstruction failures, inaccuracies, and unnatural poses, particularly under extreme masking conditions. These results highlight MaskHand’s superior capability in modeling uncertainty and synthesizing realistic meshes under severe occlusion, underscoring its robustness and practical effectiveness in challenging real-world scenarios.

\section{Impact of 2D Pose Context}
\label{sec:pose2d_est}

We investigate how the accuracy of the 2D pose estimator influences MaskHand’s 3D reconstruction performance. In our main experiments, we utilize a lightweight OpenPose estimator OpenPose due to its computational efficiency and suitability for real-time applications. To quantify how improvements in 2D pose estimation may affect overall reconstruction quality, we conduct an additional analysis comparing OpenPose predictions against ground-truth 2D keypoints on the FreiHAND dataset. As shown in Table~\ref{tab:pose2d_est}, using ground-truth 2D poses consistently yields better reconstruction metrics, notably improving both PA-MPJPE and PA-MPVPE. Although MaskHand achieves robust performance even with estimated keypoints, this analysis indicates that further advances in 2D pose estimation accuracy can directly enhance the quality of reconstructed 3D hand meshes.

\begin{table}[h!]
\centering
\vspace{-7pt}
\scalebox{1}{
\begin{tabular}{lcccc}
\toprule
\textbf{Estimator} & PA-MPJPE & PA-MPVPE & F@5mm  & F@15mm \\
\midrule
Ground Truth          & 5.2 & 5.1 & 0.834 & 0.993 \\
2D OpenPose \cite{cao2017realtime} Estimator & 5.5 & 5.4 & 0.801 & 0.991 \\
\midrule

\end{tabular}
}

\caption{Impact of 2D Pose Estimator on 3D Reconstruction Quality on Friehand dataset  \cite{zimmermann2019FreiHAND}}
\label{tab:pose2d_est}
\end{table}

\section{Confidence-Aware Unconditional Mesh Generation }
\label{sec:uncond_gen}

\begin{table}[h!]
\centering
\vspace{-7pt}
\scalebox{1}{
\begin{tabular}{lccc}
\toprule
\textbf{Methods} & APD(mm)↑  & SI(\%)↓ \\
\midrule
PCA          & 16.3 & 0.32 \\
\midrule
\textbf{MaskHand (ours)}  & \textbf{19.3}  & \textbf{0.04}    \\
\bottomrule
\end{tabular}
}

\caption{Confidence-Aware Unconditional Mesh Generation}
\label{tab:uncond-gen}
\end{table}

MaskHand enables confidence-aware unconditional 3D hand mesh generation by leveraging generative masked modeling and probabilistic sampling. We generate 2,000 hand meshes by setting the image condition to zero in the image encoder, ensuring that synthesis is driven entirely by the learned pose distribution. Using Top-100 sampling, MaskHand not only produces high-quality, diverse meshes but also quantifies the confidence of each generated hand configuration. This confidence estimation allows MaskHand to distinguish between physically plausible and invalid meshes, an advantage over diffusion-based HHMR \cite{li2024hhmr}, which lacks direct plausibility quantification.

\begin{figure}[ht]
    \centering
    \includegraphics[width=0.8\linewidth]{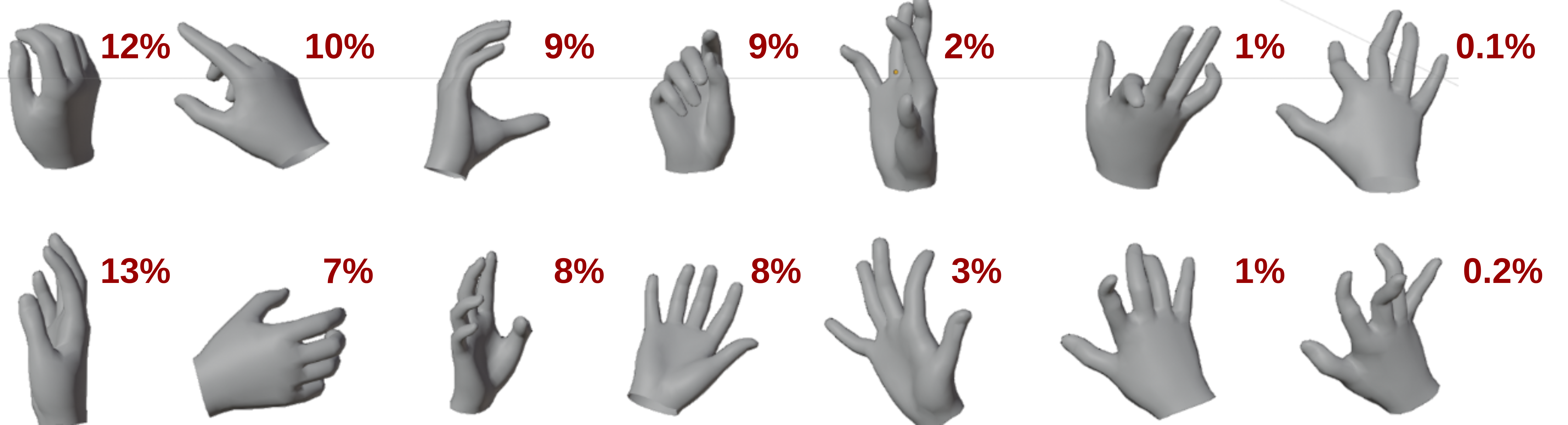}
\caption{Confidence-Aware Unconditional Mesh Generation}
    \label{fig:uncond_conf}
\end{figure}

In contrast, PCA-based generation, which samples from the MANO parameter space, produces structurally valid but limited and less diverse hand poses due to its restriction to the linear PCA subspace. The qualitative results (Figure~\ref{fig:uncond_conf}) highlight MaskHand’s ability to synthesize highly articulated hand poses, while lower-confidence samples exhibit unnatural deformations, demonstrating its uncertainty quantification capability. Quantitatively, MaskHand achieves greater diversity than PCA, as reflected in a higher APD (19.3mm vs. 16.3mm), and generates more realistic meshes, reducing self-intersection (SI) from 0.32\% to 0.04\% (Table~\ref{tab:uncond-gen}). These results confirm that MaskHand surpasses PCA in both diversity and realism, offering a principled approach to filtering implausible generations, making it a more reliable solution for unconditional 3D hand synthesis.

\section{Effectiveness of Proposed MaskHand Components}
\label{sec:comp_anal_eff}

\begin{figure}[ht] 
    \centering
    \includegraphics[width=0.9\linewidth]{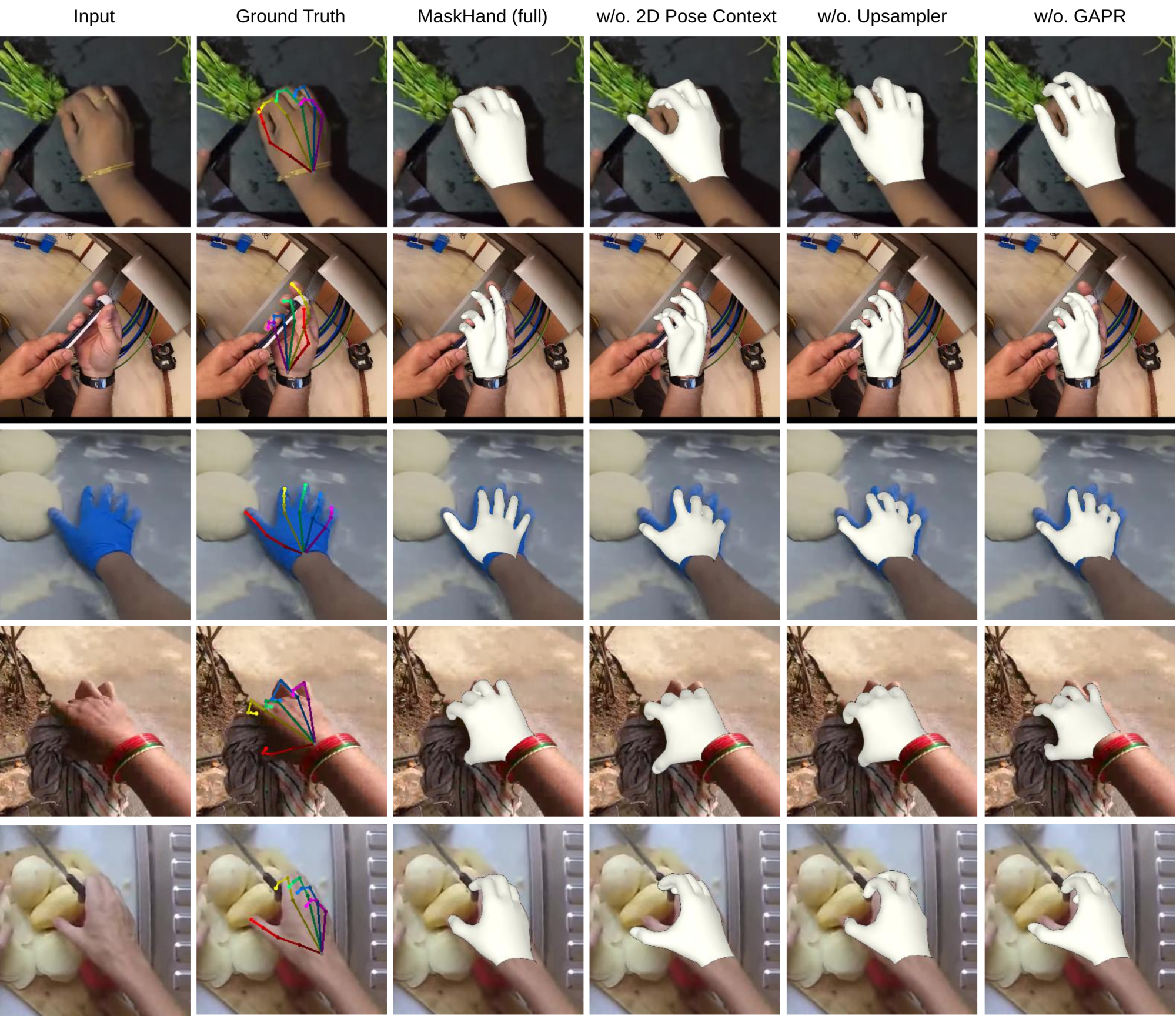}
\caption{Qualitative ablation study on component impact: Full model achieves highest accuracy, validating each component's role.}

    \label{fig:comp_analysis} 
\end{figure}

The ablation study on HO3Dv3 (Table \ref{tab:ablation_comp}) highlights GAPR as the most critical component, ensuring joint dependencies and anatomical coherence. Removing the Upsampler and 2D Pose Context slightly reduces accuracy, affecting fine details and spatial cues. The full MaskHand model achieves the best results, demonstrating the importance of these components for high-precision 3D hand mesh recovery. Qualitative comparisons in Figure \ref{fig:comp_analysis} further illustrate these effects, showing the impact of each component on reconstruction quality.

\begin{table}[h!]
\centering
\vspace{-7pt}
\scalebox{1}{
\begin{tabular}{lccccc}
\toprule
\textbf{Method} & PA-MPJPE & PA-MPVPE & F@5mm  & AUC$_J$ & AUC$_V$  \\
\midrule
w/o. Upsampler         & 7.2  & 7.2  & 0.654 & 0.857 & 0.857 \\
w/o. 2D Pose Context   & 7.1  & 7.1  & 0.656 & 0.857 & 0.858 \\
w/o. GAPR              & 7.3  & 7.3  & 0.645 & 0.853 & 0.854 \\
\midrule
\textbf{MaskHand (Full)}  & \textbf{7.0}  & \textbf{7.0}  & \textbf{0.663}  & \textbf{0.860}  & \textbf{0.860} \\
\bottomrule
\end{tabular}
}
\caption{Ablation study of testing results on the HO3Dv3 dataset \cite{hampali2022keypoint} to evaluate the impact of proposed components. 'w/o' denotes 'without'.}

\label{tab:ablation_comp}
\end{table}



\section{Masking Ratio during Training}
\label{sec:mask_ratio}

The ablation study in Table \ref{tab:masking} shows that a broader masking range \(\gamma(\tau \in \mathcal{U}(0, 0.7))\) achieves optimal results on HO3Dv3 and FreiHAND, with the lowest  PA-MPVPE values of 7.0 and 5.5, respectively. This cosine-based masking strategy, where the model learns to reconstruct from partially masked sequences, enhances robustness in 3D hand reconstruction. Narrower masking ranges, such as \(\gamma(\tau \in \mathcal{U}(0, 0.3))\), increase error, highlighting the importance of challenging the model with broader masking for better generalization.


\begin{table}[h!]
\caption{Impact of masking ratio during training on HO3Dv3 \cite{hampali2022keypoint} and FreiHAND \cite{zimmermann2019FreiHAND} datasets.}
\vspace{-7pt}
\centering
\scalebox{1}{
\begin{tabular}{ccccc}
\toprule

Masking Ratio $\gamma(\tau)$ & \multicolumn{2}{c}{HO3Dv3} & \multicolumn{2}{c}{FreiHAND} \\
\cmidrule(lr){2-3} \cmidrule(lr){4-5}
& \makecell{PA-MPJPE} & \makecell{PA-MPVPE} & \makecell{PA-MPJPE} & \makecell{PA-MPVPE} \\
\midrule
$\gamma(\tau \in \mathcal{U}(0, 0.3))$ & 7.2 & 7.2 & 6.0 & 6.1 \\
$\gamma(\tau \in \mathcal{U}(0, 0.5))$ & 7.1 & 7.1 & 5.8 & 5.7 \\
$\gamma(\tau \in \mathcal{U}(0, 0.7))$ & 7.0 & 7.0 & 5.7 & 5.5 \\
$\gamma(\tau \in \mathcal{U}(0, 1.0))$ & 7.2 & 7.3 & 6.0 & 6.0 \\
\bottomrule
\end{tabular}
}
\label{tab:masking}
\end{table}

\section{Effectiveness of Expectation-Approximated Differential Sampling}
\label{sec:diffsampling}
The results presented in Table \ref{tab:loss_impact_ho3dv2} highlight the critical role of Expectation-Approximated Differential Sampling in enabling accurate and robust 3D hand mesh recovery. The configuration utilizing all loss components—\(L_{mask}\), \(L_{MANO}\), \(L_{3D}\), \(L_{2D}\), and \(\beta\)—achieves the lowest PA-MPJPE and PA-MPVPE values of 0.70 mm on HO3Dv2 and 5.5 mm on FreiHAND, underscoring the importance of a holistic training approach. This configuration demonstrates the complementary strengths of \(L_{3D}\) in enforcing anatomical coherence, \(L_{2D}\) in mitigating monocular depth ambiguities, and \(L_{mask}\) in iteratively refining pose token predictions. Excluding critical components such as \(L_{3D}\) or \(L_{2D}\) leads to substantial degradation in performance, with errors rising to 8.1 mm on HO3Dv2 and 7.0 mm on FreiHAND. These results emphasize the necessity of these constraints for accurate 2D-to-3D alignment and plausible pose synthesis. Expectation-Approximated Differential Sampling is instrumental in this process, as it facilitates seamless integration of these losses by leveraging a differentiable framework for token refinement. This approach ensures that the latent pose space is effectively optimized, enabling the model to balance fine-grained token accuracy with global pose coherence. These findings validate the pivotal role of differential sampling in guiding the learning process, resulting in precise and confident 3D reconstructions under challenging scenarios.

\begin{table*}[h!]
\centering
\caption{Impact of different loss combinations on PA-MPJPE and PA-MPVPE errors ( in mm) for the HO3Dv2 and FreiHAND datasets.}
\vspace{5pt}
\scalebox{1}{
\renewcommand{\arraystretch}{1.3}
\begin{tabular}{lcccc}
\toprule
\textbf{Used Losses} & \multicolumn{2}{c}{\textbf{HO3Dv2}} & \multicolumn{2}{c}{\textbf{FreiHAND}} \\
\cmidrule(lr){2-3} \cmidrule(lr){4-5}
& PA-MPJPE ↓ & PA-MPVPE ↓ & PA-MPJPE ↓ & PA-MPVPE ↓ \\
\midrule
$L_{mask}, \beta$ & 7.6 & 7.5 & 6.3 & 6.4 \\
$L_{mask}, L_{MANO}, \beta$ & 7.5 & 7.5 & 6.2 & 6.4 \\
$L_{mask}, L_{MANO}, L_{3D}, L_{2D}, \beta$ & 7.0 & 7.0 & 5.7 & 5.5 \\
$L_{3D}, \beta$ & 8.1 & 7.9 & 6.8 & 7.0 \\
$L_{mask}, L_{MANO}, L_{3D}, L_{2D}, \beta$ &  7.4 & 7.3 & 6.5 & 6.7 \\
\bottomrule
\end{tabular}}
\label{tab:loss_impact_ho3dv2}
\end{table*}

\section{Confidence-Guided Masking}
\label{sec:conf_guided}
\begin{figure*}[ht] 
    \centering
    \includegraphics[width=0.9\linewidth]{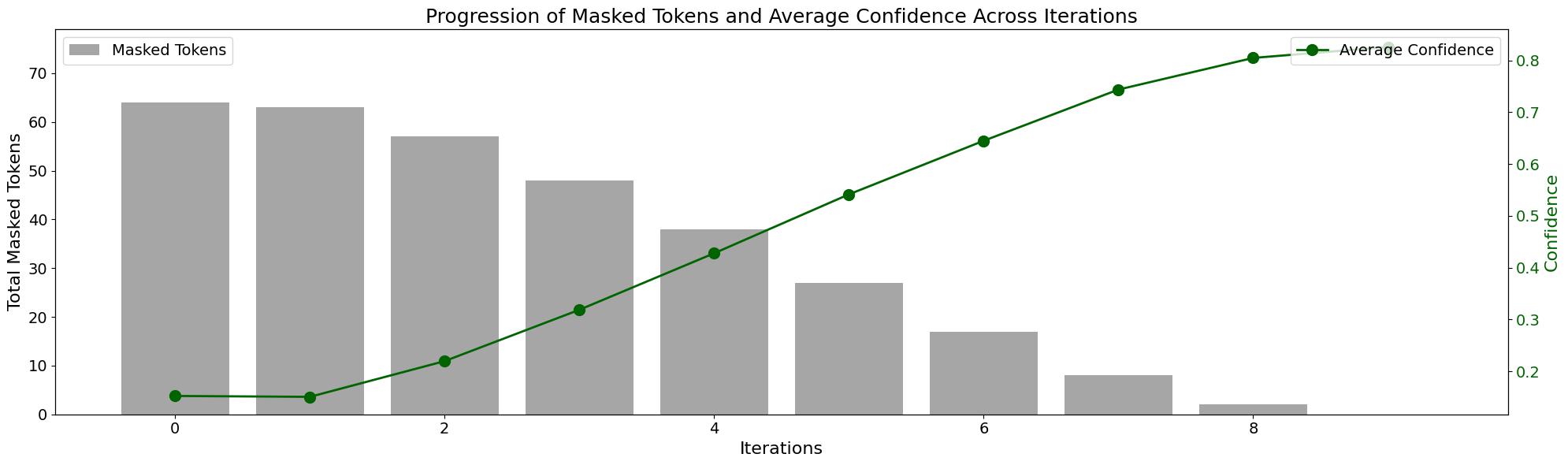}

\caption{ Progression of Masked Tokens and Average Confidence Across Iterations. This figure visualizes the iterative refinement process in Confidence-Guided Sampling. The gray bars represent the number of masked tokens at each iteration, starting from a fully masked sequence and progressively decreasing. The green curve shows the corresponding average confidence in the model's predictions, which increases steadily with iterations. This dynamic showcases the effectiveness of the sampling strategy in resolving ambiguities and refining 3D pose estimates, culminating in high-confidence predictions with minimal masking by the final iteration.}
\label{fig:MaskHand_mask_infer_conf}
\end{figure*}

Figure \ref{fig:MaskHand_mask_infer_conf} illustrates the iterative process of Confidence-Guided Sampling used during inference for refining pose predictions. The gray bars represent the total number of masked tokens across iterations, while the green line tracks the average confidence in the model's predictions. At the initial iteration, the majority of pose tokens remain masked, indicating high uncertainty. As iterations progress, the number of masked tokens decreases significantly, which aligns with a steady increase in the model's confidence. By the final iteration, only a minimal number of tokens remain masked, while the average confidence approaches its peak. This visualization highlights the systematic reduction in uncertainty and refinement of predictions over multiple iterations, enabling robust 3D pose reconstruction. 

\section{Impact of Pose Tokenizer on MaskHand }
\label{sec:mano_MaskHand}

The results presented in Table \ref{tab:stage2_cb} demonstrate the critical influence of the Pose Tokenizer's design on the performance of MaskHand. Increasing the codebook size from $1024 \times 256$ to $2048 \times 256$ yields significant improvements in both PA-MPJPE and MVE metrics across the HO3Dv3 and FreiHAND datasets. This indicates that a moderately larger codebook provides richer and more expressive pose representations, enabling better reconstruction of complex 3D hand poses. However, expanding the codebook further to $4096 \times 256$ diminishes accuracy, suggesting that an overly large codebook introduces unnecessary complexity, making it harder for the model to generalize effectively.


\begin{table}[h!]
\caption{Impact of Codebook Size (Tokens = 96) on MaskHand.}
\vspace{-7pt}
\centering
\scalebox{1}{
\begin{tabular}{ccccc}
\toprule
 & \multicolumn{2}{c}{HO3Dv3} & \multicolumn{2}{c}{FreiHAND} \\ \hline
\makecell{\# of code  $\times$ \\ code dimension} & \makecell{MPJPE \\ ($\downarrow$)} & \makecell{MVE \\ ($\downarrow$)} & \makecell{MPJPE \\ ($\downarrow$)} & \makecell{MVE \\ ($\downarrow$)} \\ 
\midrule
$1024 \times 256$ & 7.2 & 7.3 & 6.4 & 6.5  \\
$2048 \times 128$ & 7.1 & 7.2 & 6.0 & 5.9  \\
$2048 \times 256$ & 7.0 & 7.0 & 5.7 & 5.7  \\
\bottomrule
\end{tabular}
}
\label{tab:stage2_cb}
\end{table}

\section{Impact of Multi-Scale Features}
\label{sec:feats_res}

The ablation study on multi-scale feature resolutions in MaskHand (as shown in Table~\ref{tab:feat_scale_ordered}) highlights the trade-off between accuracy and computational cost. Including resolutions up to \(4\times\) yields slight accuracy gains, with PA-MPJPE reducing to 7.0 mm on HO3Dv3 and 5.7 mm on FreiHAND. However, the addition of higher resolutions, such as \(1\times\) and \(2\times\), results in inconsistent or degraded performance. Specifically, the inclusion of \(1\times, 4\times, 8\times\) scales increases PA-MPJPE to 7.2 mm on HO3Dv3 and 5.8 mm on FreiHAND. Adding \(2\times\) further worsens performance, reaching 7.6 mm on HO3Dv3 and 6.1 mm on FreiHAND, while significantly increasing computational overhead. Notably, the omission of lower-scale features (e.g., \(1\times\)) leads to performance degradation, highlighting the importance of combining fine-grained details with holistic structure. While multi-scale features remain critical, the study demonstrates that not all resolutions contribute equally, with \(1\times\) and \(4\times\) emerging as the optimal balance for accuracy and computational efficiency.

\begin{table}[h!]
\centering
\caption{Impact of feature resolutions on PA-MPJPE and PA-MPVPE errors for HO3Dv3 and FreiHAND datasets.}
\vspace{5pt}
\scalebox{1}{
\renewcommand{\arraystretch}{1.3}
\begin{tabular}{lcccc}
\toprule
\textbf{Feature Scales (Included)} & \multicolumn{2}{c}{\textbf{HO3Dv3 ↓}} & \multicolumn{2}{c}{\textbf{FreiHAND ↓}} \\
\cmidrule(lr){2-3} \cmidrule(lr){4-5}
 & \textbf{PA-MPJPE} & \textbf{PA-MPVPE} & \textbf{PA-MPJPE} & \textbf{PA-MPVPE} \\
\midrule
1$\times$ & 7.1 & 7.1 & 5.9 & 5.8 \\
1$\times$, 4$\times$ & 7.0 & 7.0 & 5.7 & 5.5 \\
1$\times$, 4$\times$, 8$\times$ & 7.2 & 7.2 & 5.8 & 5.9 \\
1$\times$, 8$\times$, 16$\times$ & 7.1 & 7.1 & 6.0 & 5.9 \\
1$\times$, 4$\times$, 8$\times$, 16$\times$ & 7.6 & 7.5 & 6.1 & 6.3 \\
\bottomrule
\end{tabular}}
\label{tab:feat_scale_ordered}
\end{table}


\section{Deformable Cross-Attention Layers in  MaskHand} 
\label{sec:deformable}
The ablation study in Table \ref{tab:freiHAND_layers} highlights the pivotal role of Deformable Cross-Attention Layers in the Context-Infused Masked Synthesizer of MaskHand. Increasing layers from 2 to 4 yields significant performance gains, reducing PA-MPJPE to 5.7 mm and PA-MPVPE to 5.5 mm on the FreiHAND dataset. This improvement underscores the layers' effectiveness in fusing multi-scale contextual features and refining token dependencies for enhanced 3D hand mesh reconstruction. However, further increasing the layers beyond 4 results in diminishing returns, with slight performance degradation at 6 and 8 layers (e.g., PA-MPVPE increases to 5.8 mm and 6.1 mm, respectively). This decline suggests that additional layers introduce unnecessary complexity, potentially overfitting or disrupting the model's ability to generalize effectively. The findings reveal that 4 layers provide the optimal balance, leveraging the benefits of cross-attention mechanisms without incurring computational overhead or accuracy trade-offs.

\begin{table}[h!]
\caption{Impact of Deformable Cross Attention Layers on FreiHAND dataset.}
\vspace{-7pt}
\centering
\scalebox{1}{
\begin{tabular}{lcccc}
\toprule
Metric & \makecell{2 Layers} & \makecell{4 Layers} & \makecell{6 Layers} & \makecell{8 Layers} \\ 
\toprule
PA-MPJPE & 6.6 & 5.7 & \textbf{5.7} & 6.0 \\
PA-MPVPE   & 6.9 & \textbf{5.5} & 5.8 & 6.1 \\
\bottomrule
\end{tabular}
}
\label{tab:freiHAND_layers}
\end{table}

\section{Qualitative Results in the Wild}
\label{sec:qualitative}

\begin{figure*}[ht] 
    \centering
    \includegraphics[width=0.9\linewidth]{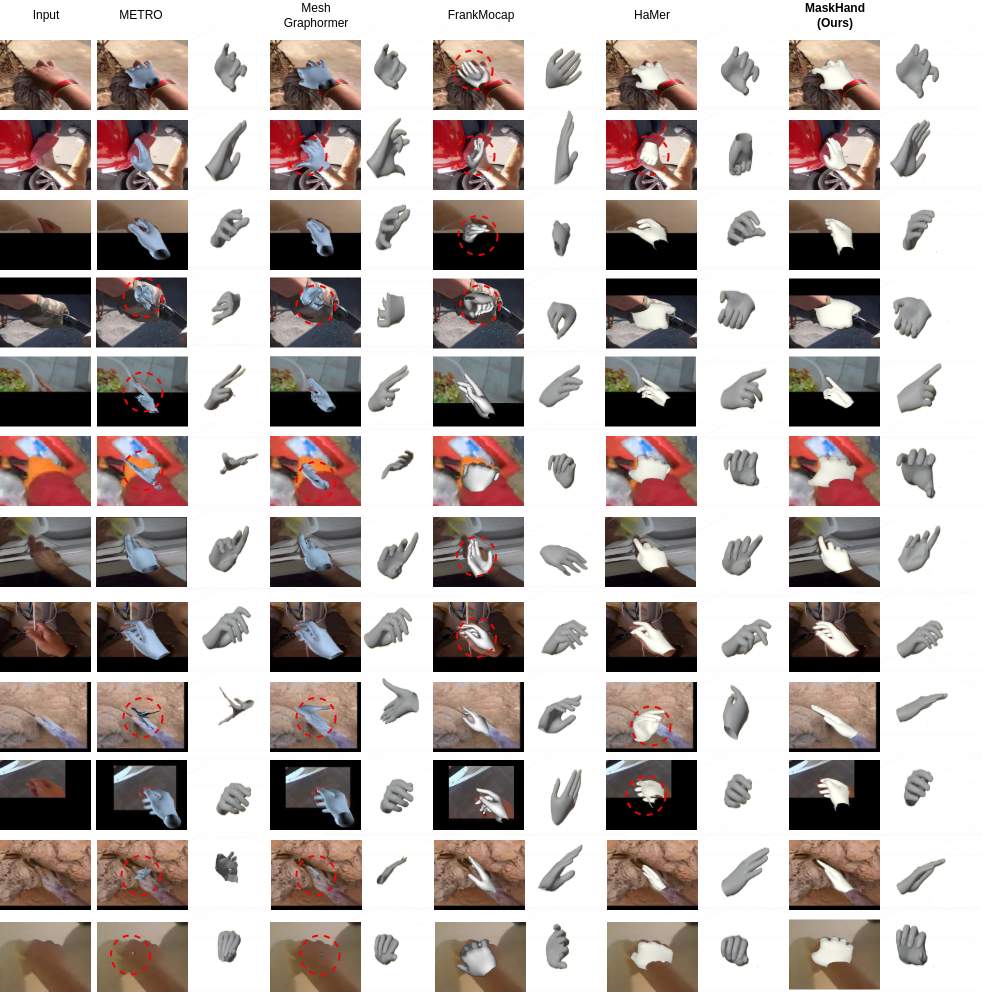}
    \caption{ Comparison of State-of-the-Art (SOTA) methods for 3D hand mesh recovery, highlighting the performance of MaskHand. Unlike other approaches, MaskHand employs a generative masked modeling framework to synthesize unobserved or occluded regions, enabling precise and robust 3D hand reconstructions even in challenging scenarios such as heavy occlusions, hand-object interactions, and diverse hand poses. This comparison underscores MaskHand's ability to outperform competing methods by addressing ambiguities in the 2D-to-3D mapping process through its innovative masked token refinement strategy.}
\label{fig:MaskHand_sota_supp}
\end{figure*}

\partitle{Comparison of State-of-the-Art (SOTA) Methods} Figure~\ref{fig:MaskHand_sota_supp} demonstrates the superiority of MaskHand over other SOTA methods in recovering 3D hand meshes. Unlike competing approaches, MaskHand employs a generative masked modeling framework, enabling it to synthesize unobserved or occluded hand regions. This capability allows MaskHand to achieve robust and precise 3D reconstructions, even in scenarios with heavy occlusions, intricate hand-object interactions, or diverse hand poses. By refining masked tokens, MaskHand effectively addresses ambiguities in the 2D-to-3D mapping process, resulting in highly accurate reconstructions.

\begin{figure*}[ht] 
    \centering
    \includegraphics[width=0.9\linewidth]{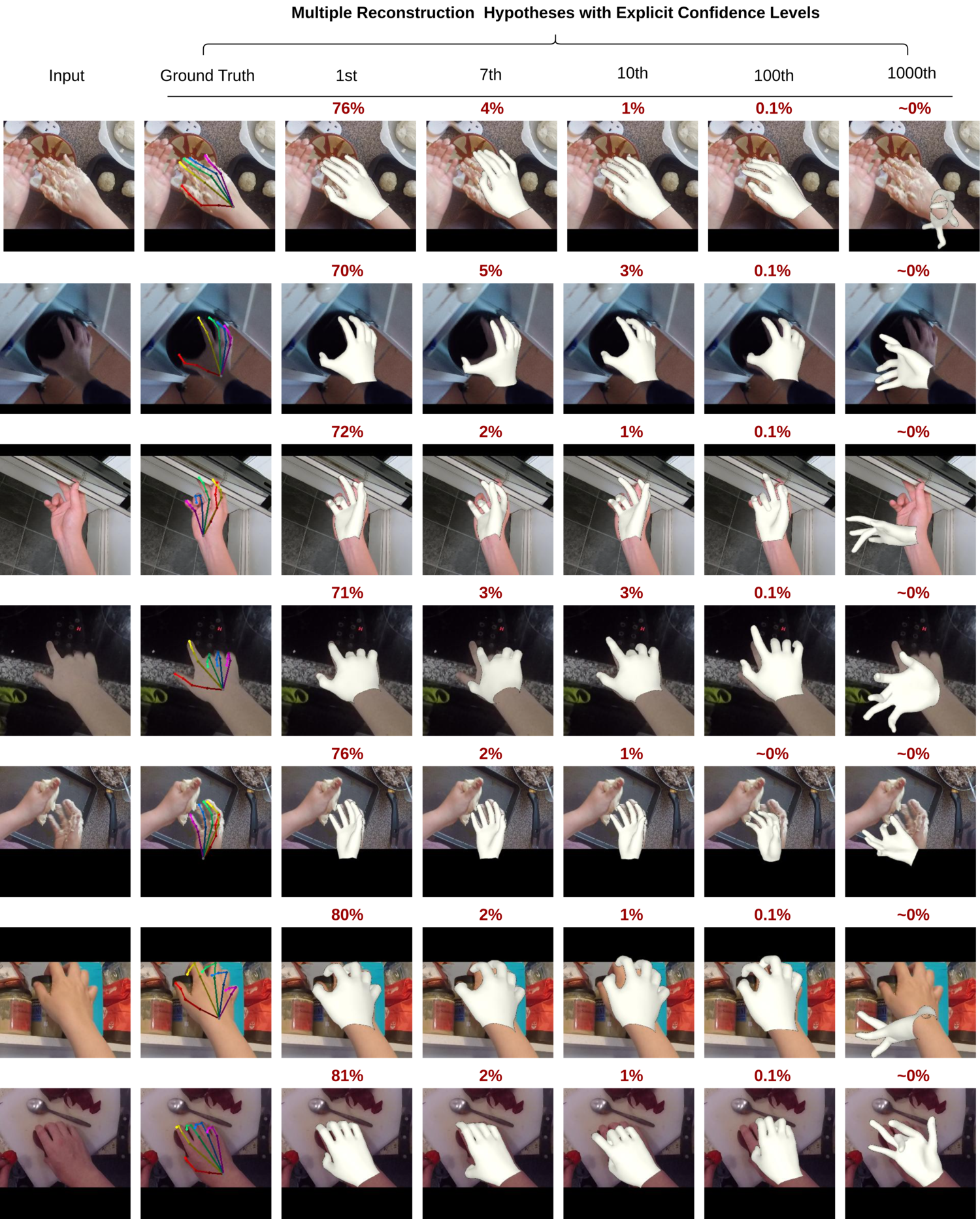}
\caption{Multiple reconstruction hypotheses with explicit confidence levels. The figure illustrates MaskHand's 3D hand mesh reconstructions in occluded scenarios, ranked by confidence. High-confidence hypotheses closely align with the ground truth, ensuring structural accuracy and fidelity. As confidence decreases (e.g., from the 100th to the 1000th hypothesis), reconstructions degrade, exhibiting distortions and unrealistic poses. This highlights the importance of prioritizing high-confidence hypotheses for robust and accurate 3D hand reconstruction.}
\label{fig:MaskHand_iters_supp}
\end{figure*}

\begin{figure*}[ht] 
    \centering
    \includegraphics[width=0.9\linewidth]{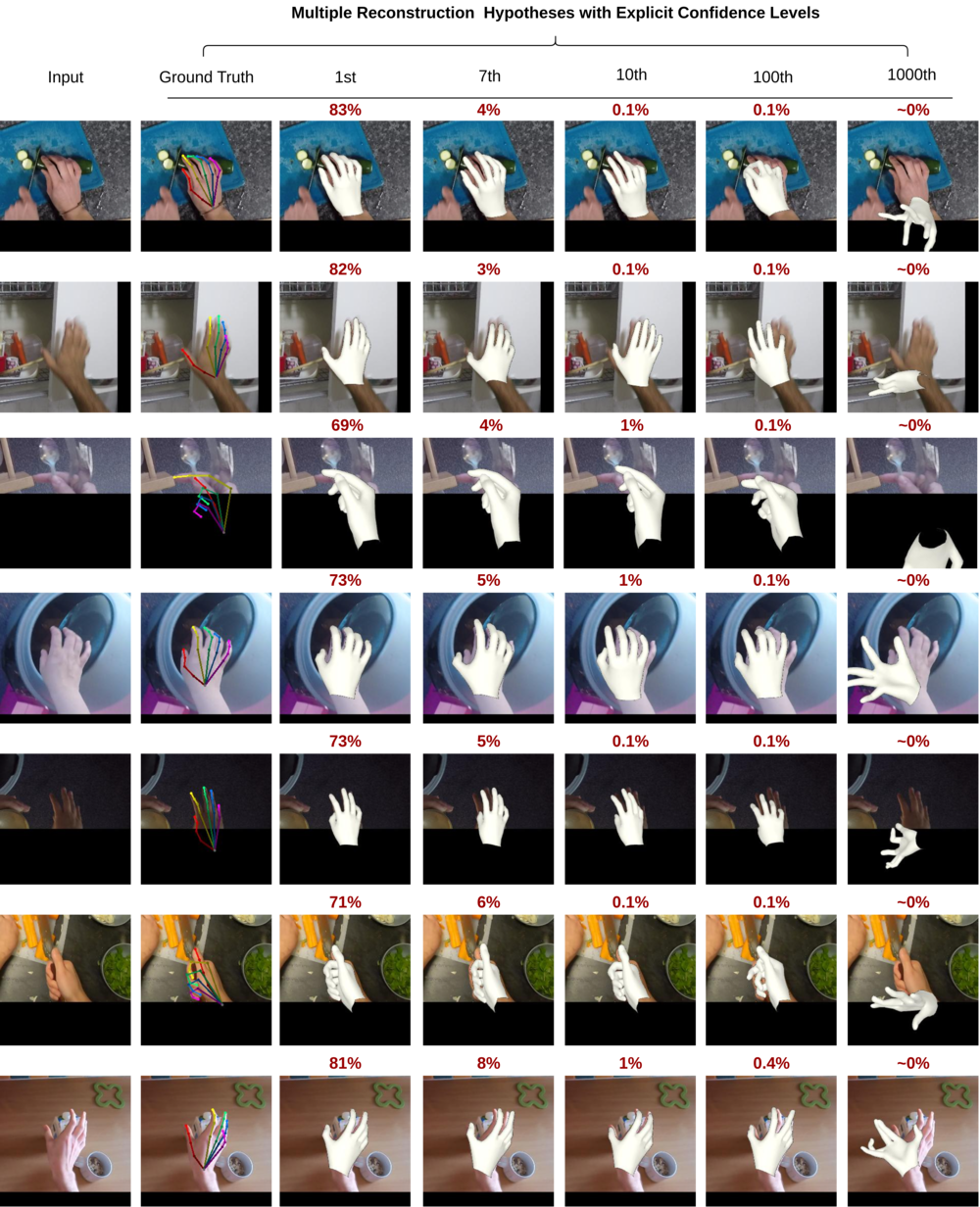}
\caption{Multiple reconstruction hypotheses with explicit confidence levels. The figure illustrates MaskHand's 3D hand mesh reconstructions in occluded scenarios, ranked by confidence. High-confidence hypotheses closely align with the ground truth, ensuring structural accuracy and fidelity. As confidence decreases (e.g., from the 100th to the 1000th hypothesis), reconstructions degrade, exhibiting distortions and unrealistic poses. This highlights the importance of prioritizing high-confidence hypotheses for robust and accurate 3D hand reconstruction.}
\label{fig:MaskHand_iters1}
\end{figure*}

\partitle{Multiple Reconstruction Hypotheses with Explicit Confidence Levels}  Figure ~\ref{fig:MaskHand_iters_supp} and ~\ref{fig:MaskHand_iters1} illustrates MaskHand's 3D hand mesh reconstructions in occluded scenarios, ranked by confidence. The comparison of reconstructions across different confidence levels reveals that high-confidence hypotheses produce meshes that closely align with the ground truth, ensuring structural accuracy and fidelity. As confidence decreases (e.g., from the 100th to the 1000th hypothesis), the reconstructions degrade, exhibiting distortions and unrealistic poses. This highlights the significance of MaskHand's confidence-aware modeling, where prioritizing high-confidence hypotheses leads to more accurate and robust 3D hand reconstructions.

\begin{figure*}[ht] 
    \centering
    \includegraphics[width=0.9\linewidth]{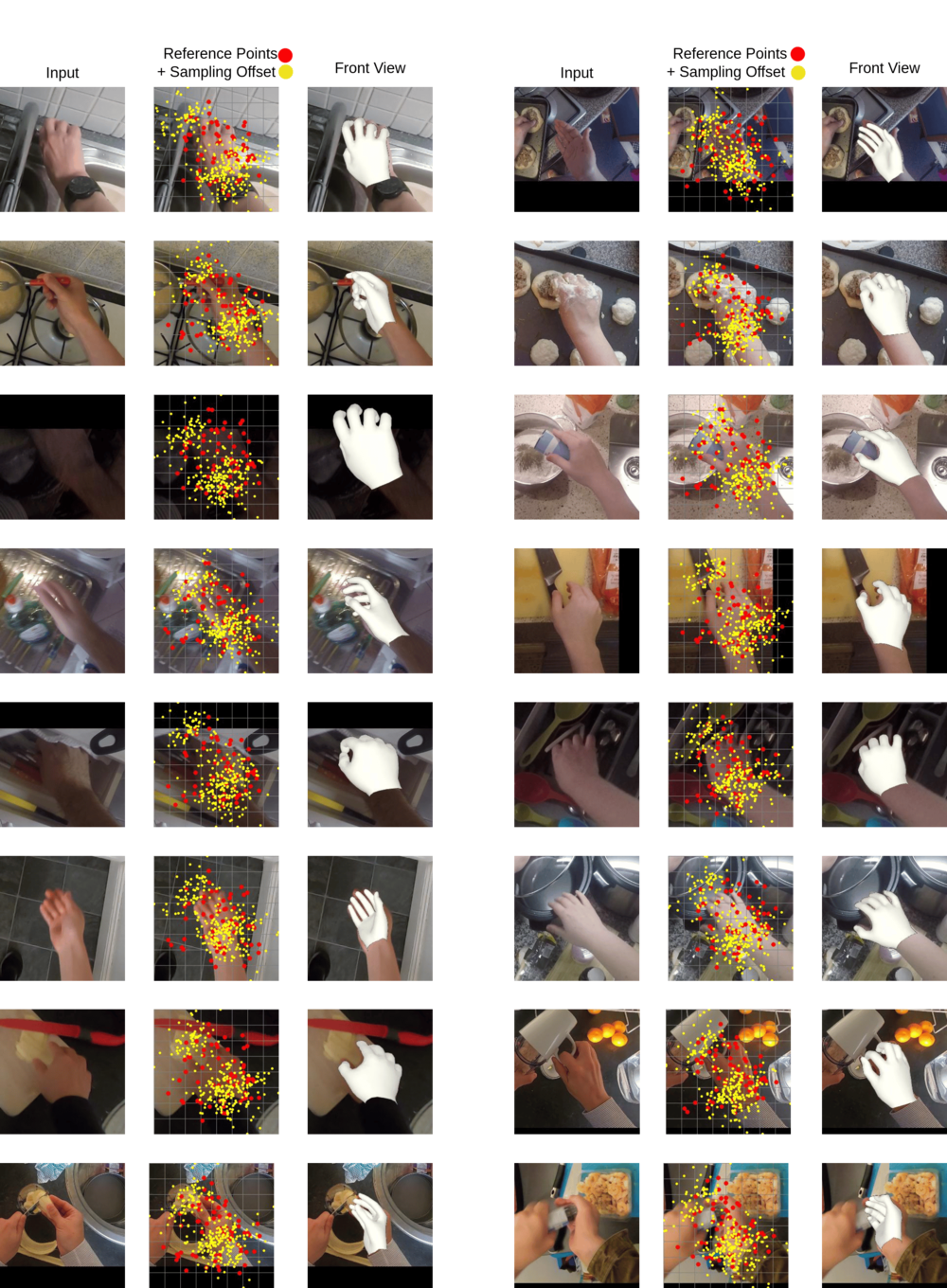}
\caption{Visualization of reference key points (yellow) and sampling offsets (red) in the Deformable Cross-Attention module of MaskHand's Masked Synthesizer. The 2D pose acts as a guidance signal, enabling the model to dynamically sample and refine features critical for reconstructing accurate 3D hand meshes (rightmost column). This mechanism adapts to challenging scenarios, such as severe occlusions, intricate hand-object interactions, and complex viewpoints, ensuring precise alignment between 2D observations and 3D predictions for robust and high-fidelity reconstructions.}
\label{fig:MaskHand_offsetsamples}
\end{figure*}

\partitle{Reference Key Points in the Deformable Cross-Attention} Figure~\ref{fig:MaskHand_offsetsamples} visualizes the interaction between reference keypoints (yellow) and sampling offsets (red) in the Deformable Cross-Attention module of MaskHand's Masked Synthesizer. By leveraging 2D pose as guidance, the model dynamically samples and refines critical features for accurate 3D reconstruction. This mechanism proves crucial in handling severe occlusions, intricate hand-object interactions, and complex viewpoints, ensuring precise alignment between 2D observations and 3D predictions.

\begin{figure*}[ht] 
    \centering
    \includegraphics[width=0.9\linewidth]{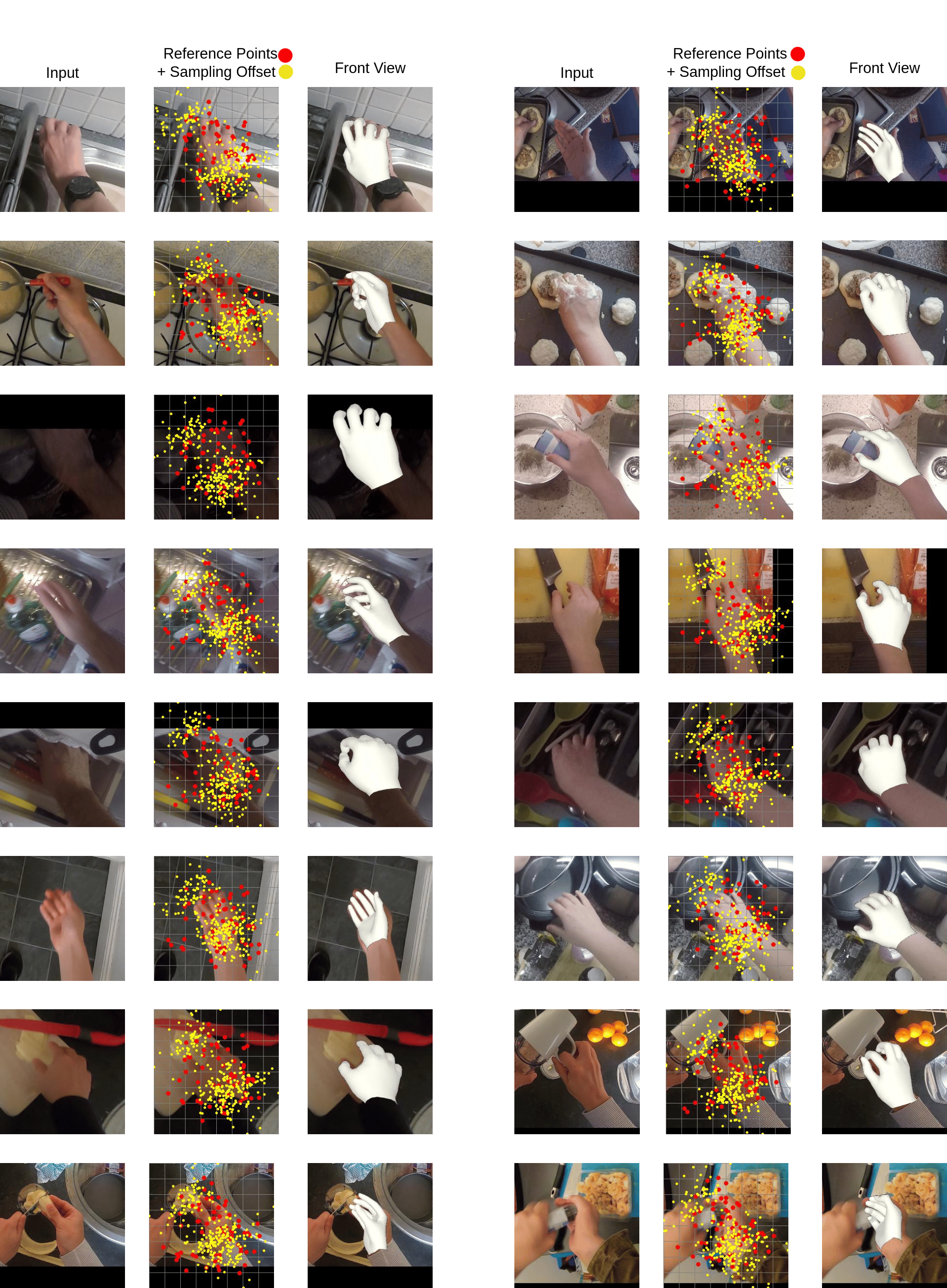}
\caption{MaskHand's performance on in-the-wild images, demonstrating its ability to recover accurate and robust 3D hand meshes from single RGB inputs. The model excels in challenging scenarios, including occlusions, hand-object interactions, and diverse hand appearances, showcasing its versatility and reliability in real-world conditions.}
\label{fig:MaskHand_inwild}
\end{figure*}

\begin{figure*}[ht] 
    \centering
    \includegraphics[width=0.9\linewidth]{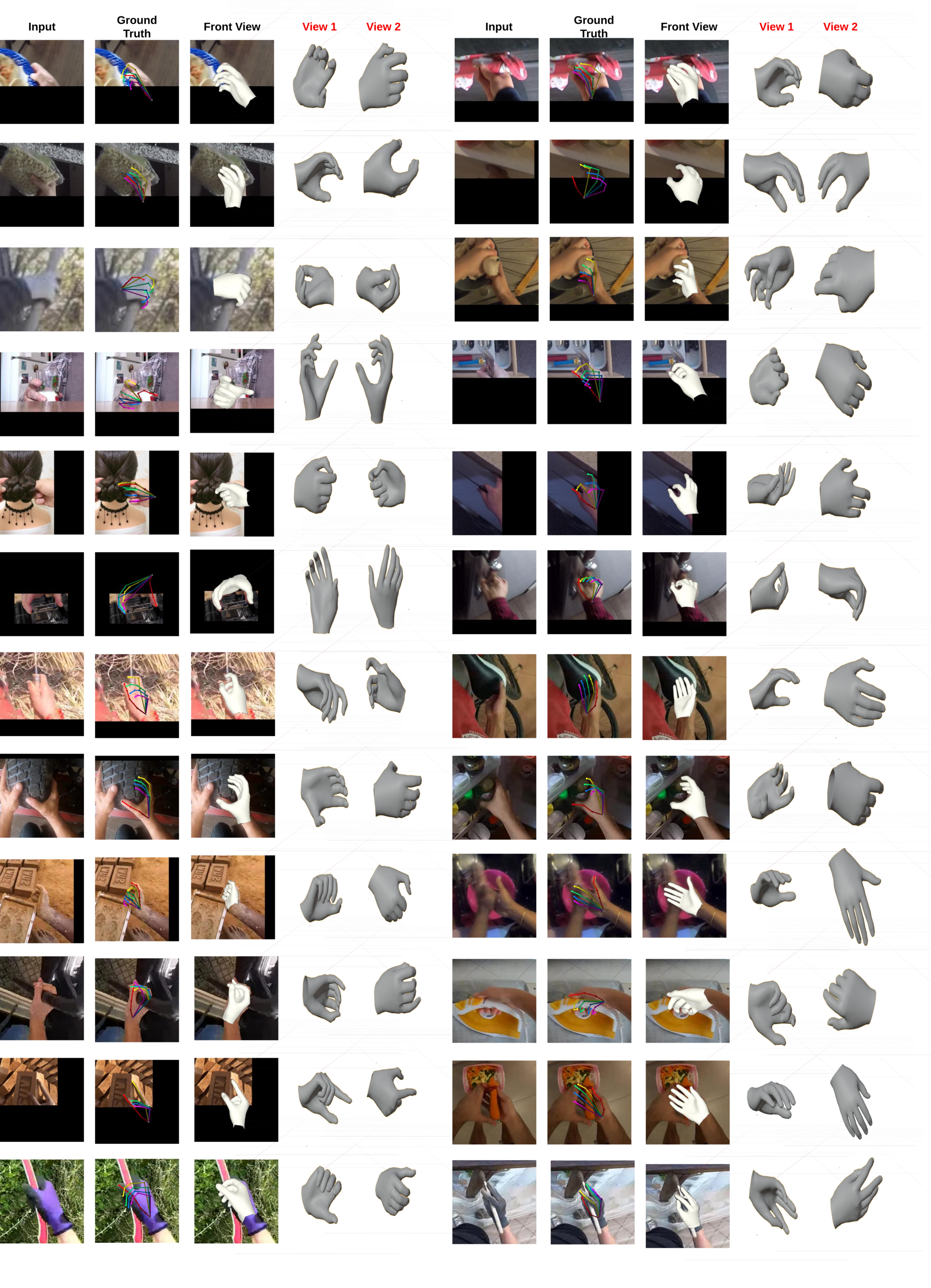}
\caption{Qualitative results of our approach on challenging poses from the HInt Benchmark \cite{pavlakos2024reconstructing} dataset.}
\label{fig:maskhand_rebuttal}
\end{figure*}

\begin{figure*}[ht] 
    \centering
    \includegraphics[width=0.9\linewidth]{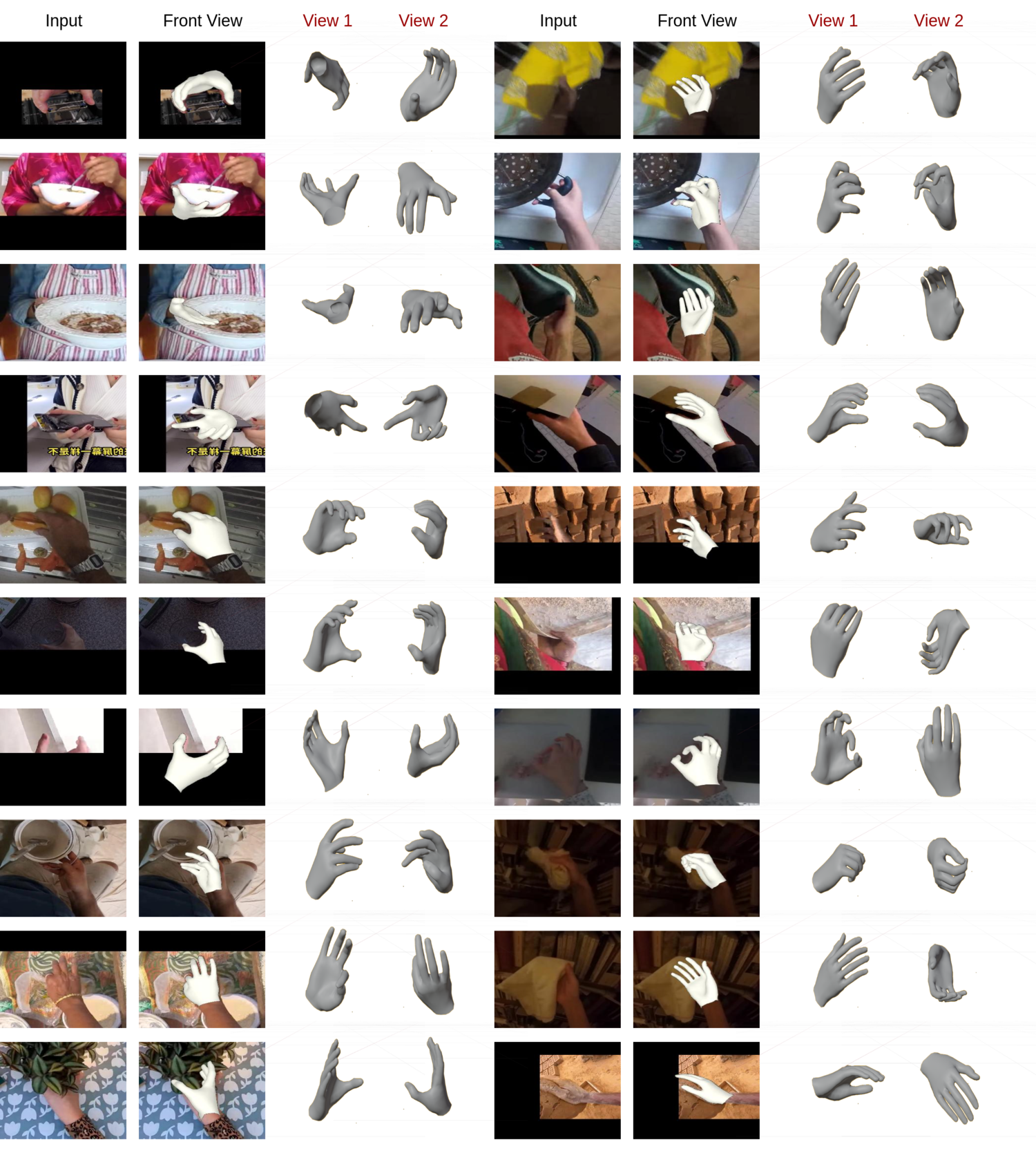}
\caption{Qualitative results of our approach on challenging poses from the HInt Benchmark \cite{pavlakos2024reconstructing} dataset.}
\label{fig:MaskHand_chall_poses_3d}
\end{figure*}

\partitle{MaskHand's Performance on In-the-Wild Images} Figure~\ref{fig:MaskHand_inwild} highlights MaskHand's robustness in real-world conditions. The model demonstrates its ability to recover accurate 3D hand meshes from single RGB images, excelling in challenging scenarios such as occlusions, hand-object interactions, and diverse hand appearances. This versatility underscores MaskHand's applicability to real-world tasks, where robust and reliable performance is essential.  
\partitle{Challenging Poses from the HInt Benchmark} Figure ~\ref{fig:maskhand_rebuttal} and  ~\ref{fig:MaskHand_chall_poses_3d} illustrates MaskHand's effectiveness in reconstructing 3D hand meshes for challenging poses from the HInt Benchmark \cite{pavlakos2024reconstructing}. The model accurately handles extreme articulations and unconventional hand configurations, showcasing its ability to generalize to complex datasets and produce high-fidelity results.

